\documentclass[sigconf, nonacm]{acmart}
\usepackage{booktabs} 

\usepackage{bbm}
\usepackage{caption}
\usepackage{enumitem}

\usepackage{pgfplots}
\usepackage{pgfplotstable}
\pgfplotsset{compat=1.18}
\usepackage{xcolor}
\usepackage[table]{xcolor}
\definecolor{best}{HTML}{FF8888}    
\definecolor{second}{HTML}{FFCCCC}  

\usepackage[ruled,vlined]{algorithm2e} 

\usepackage{pgfplots}
\usepgfplotslibrary{statistics}
\usepgfplotslibrary{groupplots}
\usepgfplotslibrary{fillbetween}
\pgfplotsset{compat=1.18}
\usepackage{xcolor}
\usepackage[export]{adjustbox}
\definecolor{myblue}{RGB}{31, 119, 180}

\SetAlFnt{\small}
\SetAlCapFnt{\small}
\SetAlCapNameFnt{\small}
\SetAlCapHSkip{0pt}

\SetAlCapHSkip{0pt}
\IncMargin{1em}
\definecolor{c_spinnerf}  {HTML}{888780}   
\definecolor{c_gscream}   {HTML}{D85A30}   
\definecolor{c_gaugroup}  {HTML}{1D9E75}   
\definecolor{c_infusion}  {HTML}{7F77DD}   
\definecolor{c_aura}      {HTML}{D4537E}   
\definecolor{c_ours}      {HTML}{185FA5}   
 
\pgfdeclarepatternformonly{hatch_ne}
  {\pgfqpoint{-1pt}{-1pt}}{\pgfqpoint{10pt}{10pt}}{\pgfqpoint{9pt}{9pt}}
  {\pgfsetlinewidth{0.4pt}\pgfpathmoveto{\pgfqpoint{0pt}{0pt}}
   \pgfpathlineto{\pgfqpoint{9pt}{9pt}}\pgfusepath{stroke}}
 
\pgfdeclarepatternformonly{hatch_nw}
  {\pgfqpoint{-1pt}{-1pt}}{\pgfqpoint{10pt}{10pt}}{\pgfqpoint{9pt}{9pt}}
  {\pgfsetlinewidth{0.4pt}\pgfpathmoveto{\pgfqpoint{9pt}{0pt}}
   \pgfpathlineto{\pgfqpoint{0pt}{9pt}}\pgfusepath{stroke}}
 
\pgfdeclarepatternformonly{hatch_cross}
  {\pgfqpoint{-1pt}{-1pt}}{\pgfqpoint{10pt}{10pt}}{\pgfqpoint{9pt}{9pt}}
  {\pgfsetlinewidth{0.4pt}
   \pgfpathmoveto{\pgfqpoint{0pt}{0pt}}\pgfpathlineto{\pgfqpoint{9pt}{9pt}}
   \pgfpathmoveto{\pgfqpoint{9pt}{0pt}}\pgfpathlineto{\pgfqpoint{0pt}{9pt}}
   \pgfusepath{stroke}}

\pgfplotsset{
  bar/.style={
    ybar,
    bar width        = 4.5pt,
    width            = \linewidth,
    height           = 5.2cm,
    enlarge x limits = 0.12,
    ymajorgrids      = true,
    grid style       = {dotted, gray!50},
    axis line style  = {gray!60},
    tick style       = {gray!60},
    xtick            = data,
    xticklabels      = {480p, 1080p, 4K},
    xtick style      = {draw=none},           
    x tick label style = {
      font      = \footnotesize,
      color     = black,
    },
    y tick label style = {
      font      = \footnotesize,
      color     = black,
    },
    title style = {
      font      = \small\bfseries,
      yshift    = -2pt,
    },
    legend style = {
      at            = {(0.5,-0.28)},
      anchor        = north,
      legend columns= 3,
      font          = \scriptsize,
      draw          = none,
      fill          = none,
      column sep    = 6pt,
      /tikz/every even column/.append style={column sep=0pt},
    },
    every axis plot/.append style={
      draw         = black!25,
      line width   = 0.3pt,
    },
    nodes near coords,
    nodes near coords align = {vertical},
    nodes near coords style = {
      font      = \tiny,
      color     = black!70,
      inner sep = 1pt,
      yshift    = 1pt,
    },
    every node near coord/.append style={
      /pgf/number format/.cd, fixed, precision=1,
    },
  },
}
 
\usepackage{pgfplots}
\usetikzlibrary{shapes}
\pgfplotsset{compat=1.18}
 
\usepackage{xcolor}
\usepackage{caption}
\usepackage{subcaption}              
 
\usepackage{booktabs}
 
\definecolor{colOne} {HTML}{4E79A7}   
\definecolor{colTwo} {HTML}{F28E2B}   
\definecolor{colFour}{HTML}{E15759}   
 
\pgfplotsset{
  resbar/.style={
    ybar,
    bar width          = 7pt,          
    width              = \linewidth,
    height             = 5.8cm,
    enlarge x limits   = {abs=20pt},
    ymajorgrids        = true,
    grid style         = {dotted, gray!45, line width=0.4pt},
    axis line style    = {gray!55, line width=0.5pt},
    tick style         = {draw=none},
    xtick              = data,
    xticklabels        = {GScream, NeRFiller, AuraFusion360, \textbf{Ours}},
    x tick label style = {
      font    = \small,
      rotate  = 25,
      anchor  = north east,
    },
    y tick label style = {font=\footnotesize},
    title style        = {font=\small\bfseries, yshift=-1pt},
    legend style = {
      at            = {(0.5, 1.18)},
      anchor        = south,
      legend columns = 3,
      font          = \small,
      draw          = gray!40,
      fill          = white,
      inner sep     = 4pt,
      column sep    = 10pt,
    },
    nodes near coords,
    nodes near coords align = {vertical},
    nodes near coords style = {
      font      = \tiny,
      color     = black!60,
      inner sep = 0.5pt,
      yshift    = 1.5pt,
    },
    every node near coord/.append style={
      /pgf/number format/.cd, fixed, precision=1,
    },
  },
}
\usepackage{xstring}
\usetikzlibrary{patterns}
\newcommand{\plotFourEighty}[1]{%
  \addplot[fill=colOne,  draw=black!15, line width=0.3pt]
    coordinates {#1};%
}
\newcommand{\plotTenEighty}[1]{%
  \addplot[fill=colTwo, draw=black!15, line width=0.3pt,
    postaction={pattern=north east lines, pattern color=white!50}]
    coordinates {#1};%
}
\newcommand{\plotFourK}[1]{%
    \StrSubstitute{#1}{+}{}[\sanitizedCoords]%
    
    \addplot[
        fill=colFour, 
        draw=black!15, 
        line width=0.3pt,
        postaction={pattern=crosshatch, pattern color=white!50}
    ]
    coordinates {\sanitizedCoords};%
}

\begin{document}
\title{3D-GIMP: When 3D Gaussian Inpainting Meets PatchMatch}

\author{Xuening Tian}
\email{Xuening.Tian@visus.uni-stuttgart.de}
\affiliation{%
  \institution{University of Stuttgart}
  \city{Stuttgart}
  \country{Germany}
}
\author{Dieter Schmalstieg}
\email{dieter.schmalstieg@visus.uni-stuttgart.de}
\affiliation{%
  \institution{University of Stuttgart}
  \city{Stuttgart}
  \country{Germany}
}

\author{Shohei Mori}
\email{shohei.mori@visus.uni-stuttgart.de}
\affiliation{%
  \institution{University of Stuttgart}
  \city{Stuttgart}
  \country{Germany}
}

\begin{abstract}
Recent advances in 3D scene editing have leveraged iterative diffusion models to update input views. However, this process is computationally expensive and struggles to produce sharp details. Meanwhile, ``hallucination drift'' frequently introduces multi-view inconsistencies, leading to structural artifacts when rendering novel viewpoints. To address this problem, we present 3D-GIMP (3D Gaussian Inpainting Meets Patch Matching), a novel hybrid paradigm designed for high-fidelity object removal in 3D Gaussian Splatting. Instead of diffusing every view, 3D-GIMP performs a single generative inpainting on a key reference view, which serves as an appearance prior. We then introduce a 3D-aware PatchMatch algorithm to propagate these reference textures across all remaining views via correspondence matching, effectively bypassing the stochastic nature of frame-by-frame diffusion. By prioritizing reconstructive consistency over iterative generation, 3D-GIMP maintains high-frequency details across arbitrary resolutions while ensuring a mathematically consistent 3D reconstruction. Our experiments demonstrate that 3D-GIMP not only achieves competitive inpainting quality as previous methods using diffusion in multiple views, but also outperforms these methods in rendering speed and view consistency.

\end{abstract}



\begin{teaserfigure}
    \centering
    \includegraphics[width=\linewidth]{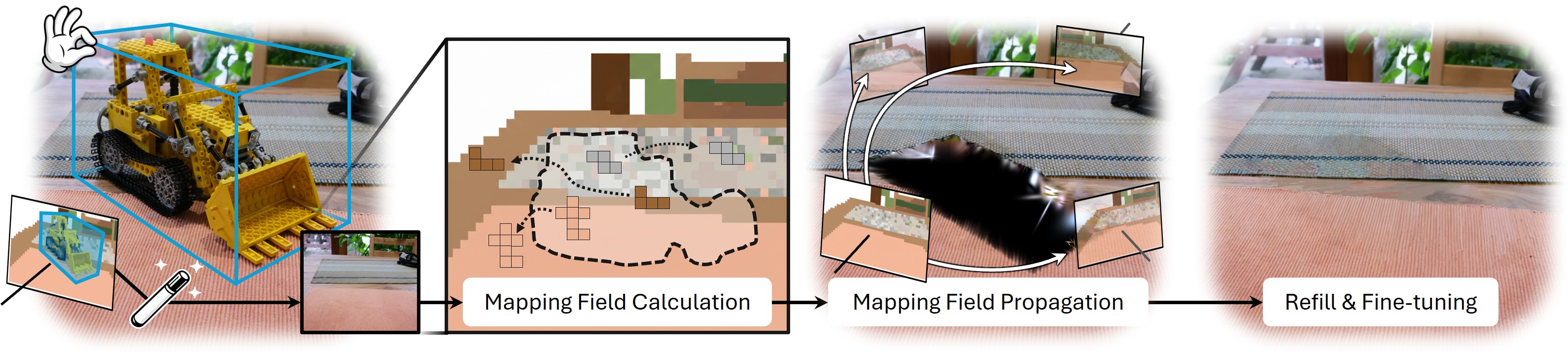}
    \caption{The proposed pipeline for view-consistent object removal in 3DGS. We initiate object removal in a reference view by defining a 3D ROI within the reconstructed Gaussian Splatting scene. Our method generates view-consistent masks and recovers missing geometry using a Poisson blending-based depth completion approach. A mapping field is then computed and propagated across following views to guide patch resampling and inpainting. Finally, through a brief optimization phase, we achieve high-quality, objectless Gaussian representations with superior efficiency.}
    \label{fig:teaser}
    \Description{teaser figure}
\end{teaserfigure}
\maketitle

\section{Introduction}

3D scene modeling from 2D images, especially  \textit{3D Gaussian Splatting} (3DGS), \cite{kerbl3Dgaussians} has achieved foundational capabilities for photorealistic novel view synthesis. Despite the popularity of 3DGS, effective editing remains a challenge. In particular, removing unwanted objects while maintaining structural and geometric coherency needs to address two primary issues: (1) ensuring that the new scene part synthesized within the disoccluded ``hole'' matches the surrounding environment with respect to surface and depth scale, and (2) ensuring that generative inpainting remains geometrically consistent when propagated across multiple training views.

\begin{figure}
    \centering
  \includegraphics[width=\linewidth]{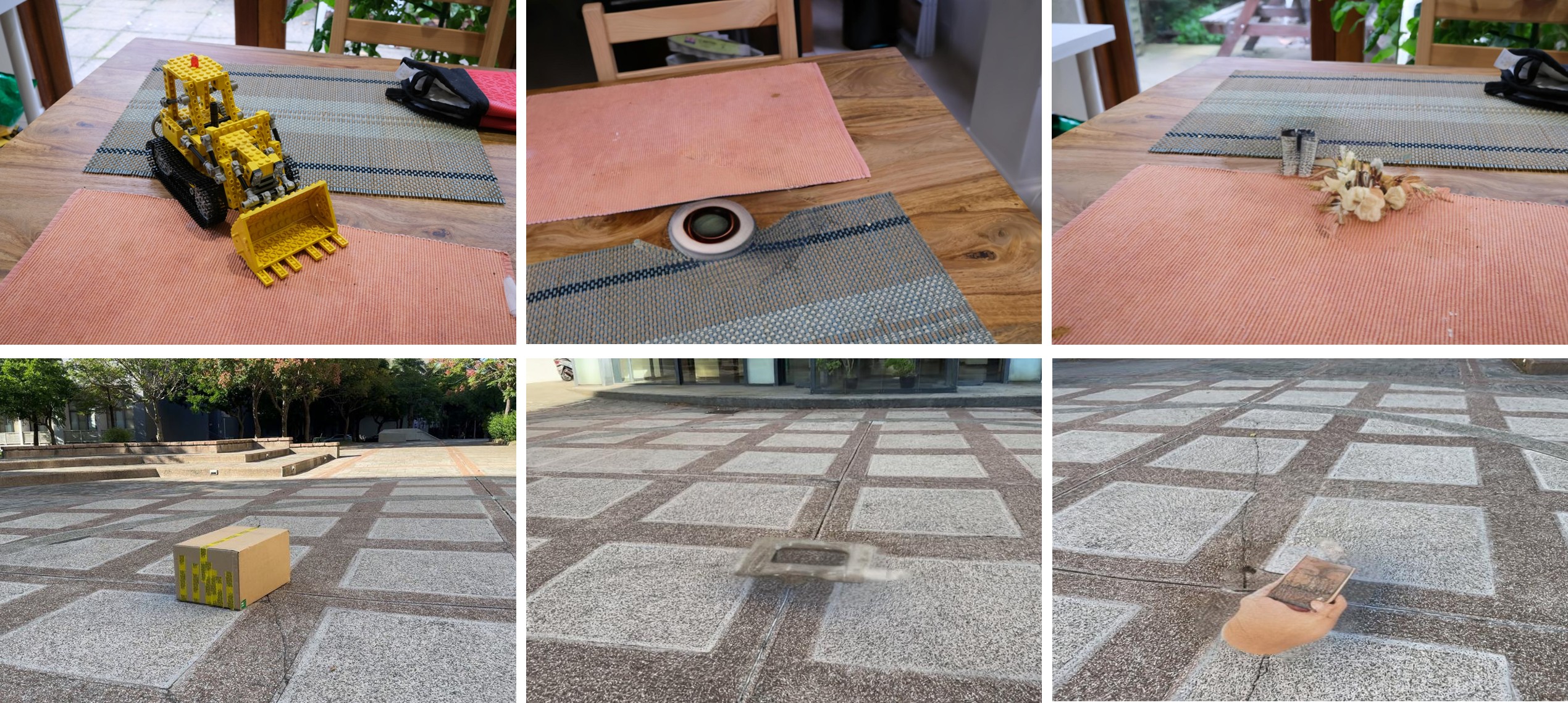}
    \caption{Random hallucinations of an image inpainting model. The original (left) and inpainted scenes (middle and right). The multi-view inconsistency degrades 3D inpainting quality (e.g., AuraFusion360 \cite{wu2025aurafusion}).}
    \Description{depth comparison}
    \label{fig:hallucinations} 
\end{figure}

Unlike NeRF \cite{nerf}, the explicit geometric primitives of 3DGS make 3D scene manipulation more straightforward, but the problem of achieving multi-view consistency between the training views remains. Inpainting the training views with diffusion models mitigates artifacts \cite{wu2025aurafusion}, but a diffusion model can produce hallucinations and requires a cumbersome conditioning of the diffusion process, which must strike a compromise between high performance and high inpainting fidelity (\figurename~\ref{fig:hallucinations}).

To address this issue, we propose multi-view consistent object removal for 3DGS inspired by the efficiency of traditional patch matching~\cite{patchmatch}. We replace the stochastic patch initialization in traditional patch matching with a stable diffusion prior and apply scale-consistent depth completion to precisely propagate to the training views. Our method enforces rigorous geometric and photometric consistency in multi-view inpainting, thereby significantly simplifying subsequent steps in scene restoration. 
Compared to diffusion-only methods, the need for further refinement is removed or at least greatly reduced. We name our method 3D Gaussian Inpainting Meets Patch Matching (3D-GIMP).

In summary, our contributions are as follows:
\begin{itemize}[leftmargin=12pt]
\item 
We propose a patch-based inpainting framework for 3DGS that establishes a global geometric basis to ensure structural integrity and high-fidelity texture synthesis in occluded regions, overcoming the limitations of traditional 2D inpainting in 3D space.
\item 
We introduce a seamless depth completion method that integrates scale-ambiguous monocular depth priors with incomplete scene depth from 3DGS by formulating the reconstruction as a gradient-domain Poisson problem.
\item 
We demonstrate that patch-based inpainting achieves performance competitive with current diffusion methods while offering significant improvements in computational efficiency and scalability. Furthermore, we show that integrating patch-matching results effectively mitigates hallucination artifacts and enhances cross-view consistency compared to purely generative approaches.
\end{itemize}
%

\section{Related work}

\subsection{Image and video inpainting}

Image inpainting using patch-matching is a well-established approach to address diverse image artifacts, including text overlays, block structures and sensor noise \cite{6960838, 7922581, 10.1007/s11042-017-4509-0}. While traditional methods primarily recover RGB data from spatial neighbors, recent advances have integrated geometric data. For example, clustering of RGB-D patches has been shown to enhance denoising performance for depth maps captured by RGB-D sensors \cite{6909829, PatchPrioritization}. Building on these foundations, video and 3D inpainting extend the challenge of reconstruction to include temporal and geometric consistency. Early approaches in this domain focused on propagating patch information through optical flow \cite{liCvpr22vInpainting} or relied on sparse point cloud projections \cite{liao2020dvi} to ensure continuity across frames.

The rise of diffusion models has shifted the focus toward generative synthesis. Methods such as Stable Diffusion XL have set new benchmarks for 2D inpainting quality by leveraging massive datasets to hallucinate missing semantic details \cite{10483967, moufad2025efficientzeroshotinpaintingdecoupled}. While these generative priors offer unprecedented texture fidelity, directly applying them to video or 3D scenes remains difficult. Without explicit conditional guidance, the stochastic nature of diffusion models often produces flickering artifacts or ``hallucinates'' the removed object back into the scene. This glitch occurs because the model naturally tries to make sense of the surrounding context; it interprets residual shadows or reflections as evidence that the object should still be there \cite{winter2024objectdrop}. The generated details shift and change across different viewpoints, creating geometric contradictions that ultimately break the realism of the 3D reconstruction.

\subsection{Neural scene representation and editing}

Reconstructing 3D spatial information from sparse or multiple camera views has long been a fundamental problem in computer vision. NeRF \cite{mildenhall2020nerf} has significantly advanced this field by implicitly representing scene radiance in its latent space. However, this implicit nature complicates scene manipulation. Remove-NeRF \cite{Weder_2023_CVPR} introduced a method for excluding views with high uncertainty, but it still struggles with blur reconstruction caused by inconsistently inpainted frames. SPIn-NeRF \cite{spinnerf} introduces a mask refinement technique that ensures consistency in 3D, but its performance remains restricted by the generative quality of the underlying 2D inpainting models.

\begin{figure*}[t]
    \centering
        \includegraphics[width=\linewidth]{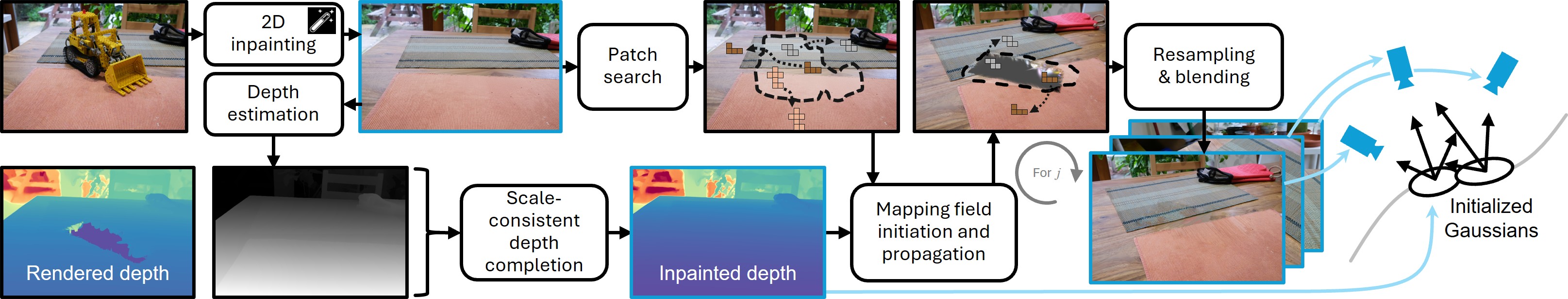}
        \caption{Data flow of our proposed pipeline. Our method selects a training view as a reference for rendering both before and after object removal. Patch searching is performed on the inpainted reference image to establish mapping correspondences. Utilizing the aligned depth map, we propagate the patch mapping field to the target views, enabling local pixel sampling for 2D inpainting. This strategy allows for effective pixel inpainting even in the presence of strong view-dependent effects, such as reflections, while using only a single view as a reference.
        }
    \label{fig:pipeline}
\end{figure*}

More recently, 3DGS has emerged as a game-changer for radiance field reconstruction, catalyzing rapid advancements in 3D object removal and scene editing \cite{chen2023gaussianeditor, wang2024gscream, imfine} because of its explicit scene representation as Gaussian primitives. Most recent object removal methods for 3DGS rely on a latent diffusion model (LDM) as a generative prior instead of a traditional convolutional neural network \cite{suvorov2021resolution}. While the Diffusion model ensures high visual fidelity in a single view, it can still exhibit random hallucinations when inpainting multiple views. The resulting inconsistencies spread across different camera angles during the inpainting process. Because the generated content does not match between views, the model cannot form a coherent 3D shape, resulting in a blurred or distorted reconstruction. We observe that any 3D inpainting method invoking a generative model struggles to repair inconsistencies with the structural prior. Our method is designed to avoid this gap by proposing a scalable and effective 3D object removal paradigm.

\section{Method}

Our work extends the ideas of InpaintFusion \cite{Mori2020InpaintFusion}, which uses traditional patch-matching for incremental 3D inpainting. InpaintFusion obtains geometry as a point cloud from RGB-D SLAM. However, InpaintFusion predates 3DGS and relies on colored surfels as a scene representation. Moreover, the copy-and-paste nature of patch-match restricts the content in the inpainted area to elements that are already present in one of the views. 

In the following sections, we introduce preliminaries (Section~\ref{subsec:preliminary}) and discuss the steps of our object removal pipeline, which are summarized in \figurename~\ref{fig:pipeline}. We remove the target object by generating object masks that are consistent across views (Section~\ref{subsec:oj_rm_and_mask}), followed by patch-matching (Section~\ref{subsec:propagation}). For filling the uncovered region with view-consistent Gaussian primitives, see the supplementary.

\subsection{Preliminaries}
\label{subsec:preliminary}

Given a 3DGS scene reconstructed from a set of training views $\mathcal{I}=\{I_j\}_{j=0}^{J}$, we define a 3D region of interest $\mathcal{R} \subset \mathbb{R}^3$ corresponding to the bounding box of an object to be removed. Our goal is to generate an inpainted 3DGS scene with synthesized content in $\mathcal{R}$ that is spatially and photometrically consistent with the surrounding in all views. 
 
\paragraph*{3D Gaussian splatting} 

3DGS represents a scene as a collection of Gaussian primitives $\mathcal{G}=\{G_i\}_{i=1}^N$. Each Gaussian is parameterized as $G_i = (\boldsymbol{\mu}_i, \Sigma_i, \mathbf{h}_i, \alpha_i)$, where $\boldsymbol{\mu}_i \in \mathbb{R}^3$ is the center of the primitive, $\Sigma_i \in \mathbb{R}^{3\times3}$ is the covariance matrix, and $\alpha_i$ is the transparency. The view-dependent RGB color
\begin{equation}
    c_i(\omega) = \sum_{k=0}^{K} h_{ik} Y_{k}(\omega)
\end{equation}
in direction $\omega$ is modeled via spherical harmonics (SH) coefficients $\mathbf{h}_i$ and SH basis functions $\mathbf{y}=[Y_k]^\top$. The final image is rendered by projecting these 3D Gaussian primitives into 2D and employing alpha-blending:
\begin{equation}
c = \sum_{i=1}^{N} c_i \alpha_i \prod_{j=1}^{i-1} (1 - \alpha_j).
\end{equation}
While 3DGS has achieved remarkable results, the underlying problem of inconsistent view-primitive intersections hinders accurate surface estimation. Therefore, \citet{Huang2DGS2024} proposed 2D Gaussian splatting (2DGS) that replaces volumetric depth sorting with Gaussian surfels and ray-splat intersection. This approach enables the extraction of highly accurate, geometrically consistent maps for depth and normals.

\paragraph*{Patch-based mapping search} 
A mapping field---aka. \textit{nearest neighbor field} in image editing~\cite{patchmatch}---defines a 2D function $\mathbf{f}: T \to \mathbb{R}^2$ that determines, for each pixel $\mathbf{p}$ in a target region $T$ of an image $I$, an offset to its most similar counterpart in the source region $I\setminus T$. Finding the optimal mapping is treated as a global minimization problem 
\begin{equation}
\min_{\mathbf{f}} \sum_{\mathbf{p} \in T} \mathcal{L}(\mathbf{p}, \mathbf{f}), 
\end{equation}
where the cost function $\mathcal{L}$ is a combination of appearance (visual similarity) and spatial matching coherence~\cite{pixmix}. To accelerate the search iteration, a propagation step and a random search procedure can be applied \cite{patchmatch}.

\subsection{Object removal and multi-view-consistent masks}
\label{subsec:oj_rm_and_mask}

To accurately remove Gaussian primitives associated with a target object in a 3D scene requires identifying the object in each of the views. Most existing work either uses pre-defined 2D object masks \cite{you2025instainpaint, spinnerf} or employs a mask refinement step \cite{wu2025aurafusion} based on a dedicated 2D segmentation model such as SAM \cite{sam}. However, relying on 2D masks without knowing the shape of the corresponding 3D object is inherently prone to multi-view inconsistency. The ROI boundary is prone to shifts from view to view, making it difficult to maintain stable 3D geometry during inpainting.

Instead, we define the \textit{region of interest} (ROI) as a 3D bounding box $\mathcal{R}\in\mathbb{R}^6$ that contains the object to be removed. This approach yields an unambiguous boundary that remains perfectly consistent in all viewpoints. Specifying a 3D bounding box is intuitive for users and can be performed by annotating a 3D volume, selecting semantic Gaussian clusters \cite{gaussian_grouping}, or indirectly by applying semantic segmentation and then growing a 3D bounding box around the segments.

Using $\mathcal{R}$, the inpainting mask is established with the following steps: (1) We project $\mathcal{R}$ to the image. (2) We enlarge the projected region in 2D by a scaling parameter $s$ to create a ``guard band''. We empirically chose $s=1.2$, as it needs to be larger than the projected ROI to have a clear removal boundary. (3) We select all Gaussian primitives that have their center inside the guard band. (4) We subtract the aggregated 2D footprint of the selected Gaussian primitives from the original 2D projection of $\mathcal{R}$. The resulting target region $T$ is consistent in all views.

\subsection{Patch matching and propagation}
\label{subsec:propagation}

We begin our method by defining a reference view that has been pre-inpainted using an off-the-shelf model (the reference view selection strategy is detailed in the Suppl.). Following prior work \cite{imfine, wu2025aurafusion}, warping the reference view directly to target views is an intuitive strategy for maintaining multi-view consistency. However, this naïve warping often introduces distorted textures and structural mismatches with neighboring pixels, primarily because the estimated depth is not scale-consistent across views, and pixel colors often shift due to view-dependent effects \cite{imfine}. To address these artifacts, we introduce a Poisson-blending-based depth completion method paired with patch-match-based inpainting refinement.

\paragraph*{Scale-consistent depth completion}
\label{subsubsec: depth comple}

To fill in the incomplete depth map during Gaussian inpainting, some recent work \cite{imfine, wang2024gscream} adopts zero-shot depth estimation models. However, monocular depth estimation models produce locally consistent, but globally inconsistent scales. Other methods use an adaptive depth diffusion model \cite{wu2025aurafusion}, but still fail to complete depth on complicated geometry. Therefore, we propose an alternative depth completion method that ensures scale consistency by solving a Poisson equation.
To solve for the optimal disparity $d$, we discretize the Poisson equation on the pixel grid. Let $\mathbf{p}$ be a pixel in the target region $T$, and let $N(\mathbf{p})$ denote its set of four-connected neighbors. We define the discrete Laplacian operator as $\Delta d(\mathbf{p}) = \sum_{\mathbf{q} \in N(\mathbf{p})} (d(\mathbf{q}) - d(\mathbf{p}))$. To align the monocular depth $d_m$ with the scene scale, we incorporate a scale parameter $s_f$ into the guidance field. The discrete Poisson equation is formulated as:
\begin{equation}
\Delta d(\mathbf{p}) = \text{div}(\mathbf{v}(\mathbf{p})), \quad \forall \mathbf{p} \in T,
\end{equation}where the guidance divergence $\text{div}(\mathbf{v}(\mathbf{p}))$ is derived from the scaled monocular disparity:
\begin{equation}\text{div}(\mathbf{v}(\mathbf{p})) = \sum_{\mathbf{q} \in N(\mathbf{p})} s_f \left( d_m(\mathbf{p}) - d_m(\mathbf{q}) \right).\end{equation}
To ensure a seamless integration with the existing scene, we enforce a Dirichlet boundary condition on the boundary $\partial T$ of the target region:
\begin{equation}d(\mathbf{q}) = d_g(\mathbf{q}), \quad \forall \mathbf{q} \in \partial T,
\end{equation}
where $d_g$ represents the disparity rendered from the rasterizer.

\paragraph*{Mapping field initiation and propagation}

Classic inpainting with patch-matching~\cite{exemplar-inpainting} aims to find the most similar patches from the source region. Due to the limited pixel information retrieved from the boundary of the hole, the matching result is always suboptimal, leading to lower visual fidelity. Inspired by image retargeting, we propose a patch-matching and propagation strategy that combines prior knowledge of large foundation models~\cite{peng2024noisenuanceadvancesdeep} with local features in the target image.

\begin{algorithm}[t]
\caption{Cross-View Patch Inpainting}
\label{alg:patch inpainting}
\KwIn{Reference view $\mathcal{I}_0$, target views $\{\mathcal{I}_t\}$, target region $T_0$, mapping field $\mathbf{f}$}
\KwOut{Inpainted training set $\{\mathcal{I}_t^*\}$}
\BlankLine
\ForEach{target view $\mathcal{I}_t \in \{\mathcal{I}_t\}$}{
    \ForEach{pixel $\mathbf{p}_0 \in T_0$}{
        $\mathbf{p}_t \leftarrow \textsc{Warp}\!\left(\mathbf{p}_0,\, \pi_0,\, \pi_t\right)$\;
        $\mathbf{q}_0 \leftarrow \mathbf{p}_0 + \mathbf{f}(\mathbf{p}_0)$\;
        $\mathbf{p}_t^{\,\mathrm{tgt}} \leftarrow \textsc{Warp}\!\left(\mathbf{q}_0,\, \pi_0,\, \pi_t\right)$\;
        \If{$\mathbf{p}_t \notin \mathcal{I}_t$ \text{ or } $\mathbf{p}_t^{\,\mathrm{tgt}} \notin \mathcal{I}_t$}{
            $\left(\mathbf{p}_t,\, \mathbf{p}_t^{\,\mathrm{tgt}}\right) \leftarrow \textsc{ReTargeting}\!\left(\mathbf{p}_t,\, \mathbf{p}_t^{\,\mathrm{tgt}},\, \mathcal{I}_0,\, \mathcal{I}_t\right)$\;
        }
    }
    $\mathcal{C}_{\mathrm{cand}} \leftarrow \textsc{PixelSampling}\!\left(\mathcal{I}_t,\, \{\mathbf{p}_t, \mathbf{p}_t^{\,\mathrm{tgt}}\}\right)$\;
    \vspace{4pt}
    $\mathcal{I}_{t}^{\mathrm{raw}} \leftarrow \textsc{SortingAndBlending}\!\left(\{\mathbf{p}_t\},\, \mathcal{C}_{\mathrm{cand}}\right)$\;
    \vspace{4pt}
    $\mathcal{I}_t^* \leftarrow \textsc{PoissonBlending}\!\left(\mathcal{I}_t,\, \mathcal{I}_{t}^{\mathrm{raw}}\right)$\;
    \vspace{4pt}
    $\textsc{UpdateCameraState}\!\left(\mathcal{I}_t,\, \mathcal{I}_t^*\right)$\;
}
\end{algorithm}

\begin{figure}[t]
  \centering
  \includegraphics[width=\linewidth]{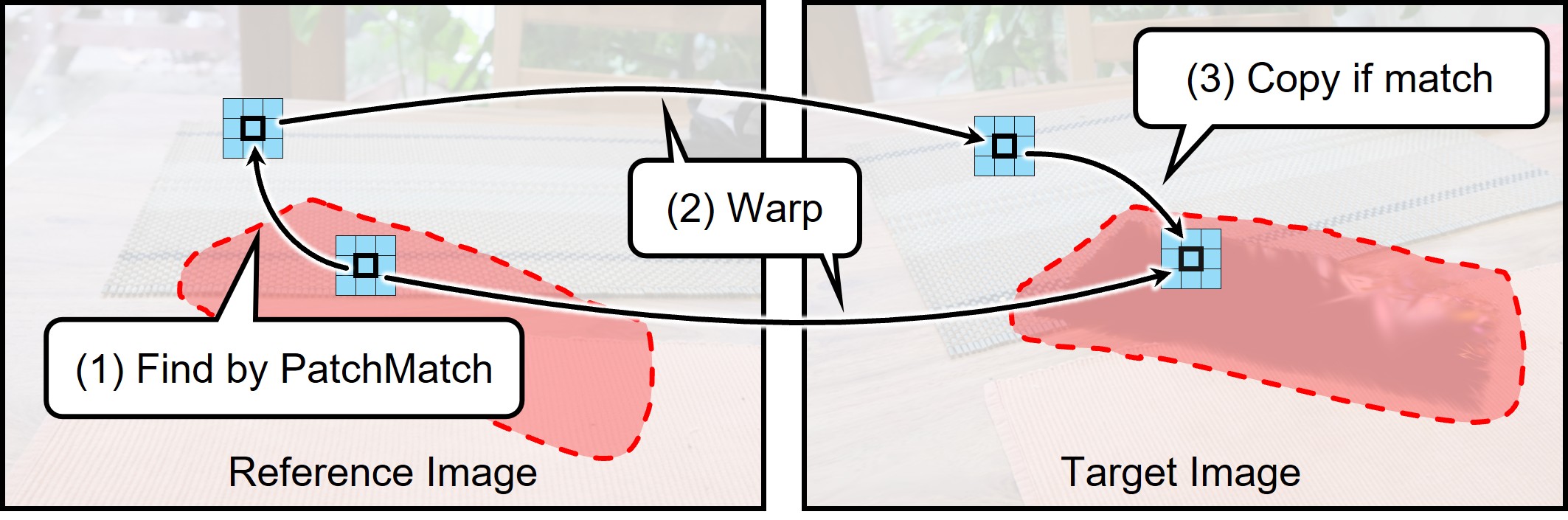}
  \caption{Illustration of the process of warping the patch matching field.}
  \label{fig:progation}
\end{figure}

As shown in \figurename~\ref{fig:progation}, after inpainting a region $T_0$ in a single reference view $I_0\in\mathcal{I}$, we construct a patch mapping field $\mathbf{f}$ by selecting a patch center $\mathbf{p} \in T_0$ and searching for a similar patch with center $\mathbf{q} \in I_0 \setminus T_0$, so that $\mathbf{q}=\mathbf{p}+\mathbf{f}(\mathbf{p})$ and $\sum_{\delta} |I_0(\mathbf{p}+\delta)-I_0(\mathbf{f}(\mathbf{p})+\delta)|<\epsilon$ for a threshold $\epsilon$ and an offset $\delta$ in a patch around $\mathbf{p}$. 

The field $\mathbf{f}$ explains the inpainted ROI in terms of observations in the non-inpainted part of the reference view. Local surface patches typically exhibit high photometric correlation in neighboring views. Therefore, we project matching pairs $(\mathbf{p}, \mathbf{q})$ to other views $I_t$ with $t>0$ using the reprojection $\mathbf{w}_t$ from $I_0$ to $I_t$. Let $\mathbf{K}_t$ be the intrinsic matrix of $I_t$, and, $[\mathbf{R}_t|\mathbf{t}_t]$, the transformation from $I_0$ to $I_t$. A point $\mathbf{p}=[u,v]^\top$ in $I_0$ is reprojected by
%
$\mathbf{w}_j(\mathbf{p})=\pi(\mathbf{K}_j[\mathbf{R}|\mathbf{t}][D(u,v)\mathbf{K}_0^{-1}\tilde{\mathbf{p}} | 1]^\top)$,
%
where $D$ is the completed reference depth map, $\tilde{\mathbf{p}}=[u,v,1]^\top$ and $\pi([x,y,z]^\top)=[x/z,y/z]^\top$.
Details of this process for cross-view patch-based inpainting are shown in Algorithm \ref{alg:patch inpainting}. 

Ideally, we wish to construct the patch mapping field $\mathbf{f}$ only once in the reference view and reuse these assignments in all reference views $I_j$ by reprojecting the mapping field. However, this approach is too rigid in practice. Source pixels are frequently occluded or off-screen in neighboring views, making it necessary to search through a local neighborhood to recover valid textures. To minimize these frequent re-mappings, we define a visibility score $v(\mathbf{p})$ for each pixel $\mathbf{p}\in T_0$. The visibility score is defined as the fraction of all views in which the pixel remains visible, denoted as 
\begin{equation}
    v(\mathbf{q}) = \frac{1}{N} \sum_{j=1}^{N} \mathbbm{1}_{I_j} \left(\mathbf{w}_j (\mathbf{q})\right),
\end{equation}
where $\mathbbm{1}$ is the indicator function. We incorporate this score as a soft constraint within the patch-match cost function to search in the reference frame. This encourages the stochastic patch-match search to prioritize pixels with high multi-view visibility, ensuring the mapping remains robust even under significant viewpoint changes. The cost function is given as 
\begin{equation}
    \mathcal{L}(\mathbf{p}, \mathbf{q}) = (1-\lambda_s) \cdot \mathcal{L}_{a}(\mathbf{p}, \mathbf{q}) + \lambda_s \cdot \mathcal{L}_{s}(\mathbf{p}, \mathbf{q})+\lambda_v\cdot v(\mathbf{q})
\end{equation}
where a spatial term $\mathcal{L}_s$ weighted by $\lambda_s$ enforces clustering of neighboring pixels, and the visibility score weighted by $\lambda_v$ encourages high visibility. We use this cost function to determine the optimal mapping field
\begin{equation}
\mathbf{f}^*=\arg\min_{\mathbf{f}} \sum_{\mathbf{p}\in T_0}
  \mathcal{L}\left(\mathbf{p},\mathbf{p}+\mathbf{f}(\mathbf{p})\right).
\end{equation}
To handle many-to-one projections and occlusions in the forward mapping, we implement a dense patch-based splatting and pixel-wise depth test. This approach ensures that occlusions are correctly resolved by retaining only the fragment closest to the center of the target camera.

With the mapping field established, the input views are inpainted by sampling and blending patches from the mapping field. We then create new Gaussians through depth-based backprojection. These consistent views then serve as the basis for 3DGS reconstruction, where Gaussian primitives in the ROI are further optimized for spatial and textural accuracy.

\section{Experiments}

We conduct qualitative and quantitative comparisons with previous methods on 360° object removal tasks using several benchmarks.

\subsection{Implementation}
\label{subsec: implementation}
We implement our pipeline by extending 2DGS \cite{Huang2DGS2024}, which supports a more accurate geometry than the original 3DGS. In addition, we also provide comparisons with the original 3DGS implementation of \citet{kerbl3Dgaussians}. The reference view is inpainted by Nano Banana Pro with Gemini 2.5~\cite{nanobanana2025}.

To ensure fair comparison, we use the first ground truth frame from the testing set as the reference view $I_0$, same for the baseline method if a reference camera is needed. We use $\lambda_s=0.001$ to balance between the neighbour pixel continuity and appearance similarity. We used patches of $5\times5$ pixels and alpha blending for overlapping pixels. All of our experiments are conducted on a single NVIDIA RTX 5090 GPU.

\subsection{Experimental setup}

\paragraph*{Dataset}

We evaluate our method on three 360° unbounded scene datasets: IMFine \cite{imfine}, 360-USID \cite{wu2025aurafusion}, and Mip-NeRF 360 \cite{multinerf2022}. These 360° scenes generally present a more challenging case than forward-facing scenes, such as those in the widely used SPInNeR dataset \cite{spinnerf}. The 360-USID dataset contains seven unbounded scenes at $960 \times 540$ resolution. Each scene provides approximately 200 views containing a target object, 30 ground-truth views of the background alone, and one reference view for inpainting guidance. IMFine includes 20 diverse scenes with highest resolution of 1920$\times$1080 in a similar setup without a reference view. Following similar quality and characteristics of 360-USID and excluding scenes with a similar background, we select eight scenes for the evaluation, including four indoor (bin, bucket, dabao, desk3) and four outdoor scenes (detergent, sofa, msi, rocks). Similarly to the previous work \cite{imfine}, we use half-resolution on IMFine for all the following evaluations and utilize the same reference view for different methods.

\paragraph*{Metrics}
We evaluate our results on four metrics: PSNR, Learned Perceptual Image Patch Similarity (LPIPS) \cite{lpips}, Frechet Inception Distance (FID) \cite{fid}, and computation time. Note that the PSNR and LPIPS values reported in our paper are all computed on the object mask. We only consider the time spent on inpainting, not the 3DGS (re-)optimization.

\subsection{Comparison with baseline and state of the art}
\label{sect:4.3}
\paragraph*{Quantitative evaluation}

We compare our method with several baselines \cite{weber2023nerfiller, spinnerf, wang2024gscream, liu2024infusion} on the IMFine dataset and 360-USID dataset from recent AuraFusion360 \cite{wu2025aurafusion}. The results are summarized in Table \ref{table:imfine}. 

InFusion achieves the lowest inference time, but only if the time it spends on depth inpainting is excluded from the measured time. Moreover, the overall quality of InFusion is significantly affected by its inaccurate depth completion, similar to AuraFusion360, leading to visible artifacts. While IMFine achieves competitive PSNR, it requires a prohibitive one-hour optimization per scene. In contrast, our 2DGS-based pipeline achieves comparable—and in the case of FID, superior—visual fidelity ($19.09$ vs $19.67$ PSNR) while reducing the computational overhead by 90\% (6 minutes). This positions our method as the first truly practical solution for high-quality, view-consistent object removal in time-sensitive applications.

\paragraph*{Qualitative comparisons}
We perform qualitative comparisons using the previous IMFine dataset in Figure \ref{fig/imfine_render.png}, but also a combination of scenes from the Mip-NeRF 360 dataset \cite{barron2022mipnerf360} and the ``bear'' scene from Instruct-NeRF2NeRF \cite{instructnerf2023}.

\begin{figure}[t]
  \centering
\begin{minipage}{\linewidth}
    \hspace{0.50\linewidth} \small Ours 
    \hspace{0.13\linewidth} \small AuraFusion360 
  \end{minipage} \\
  \includegraphics[width=\linewidth]{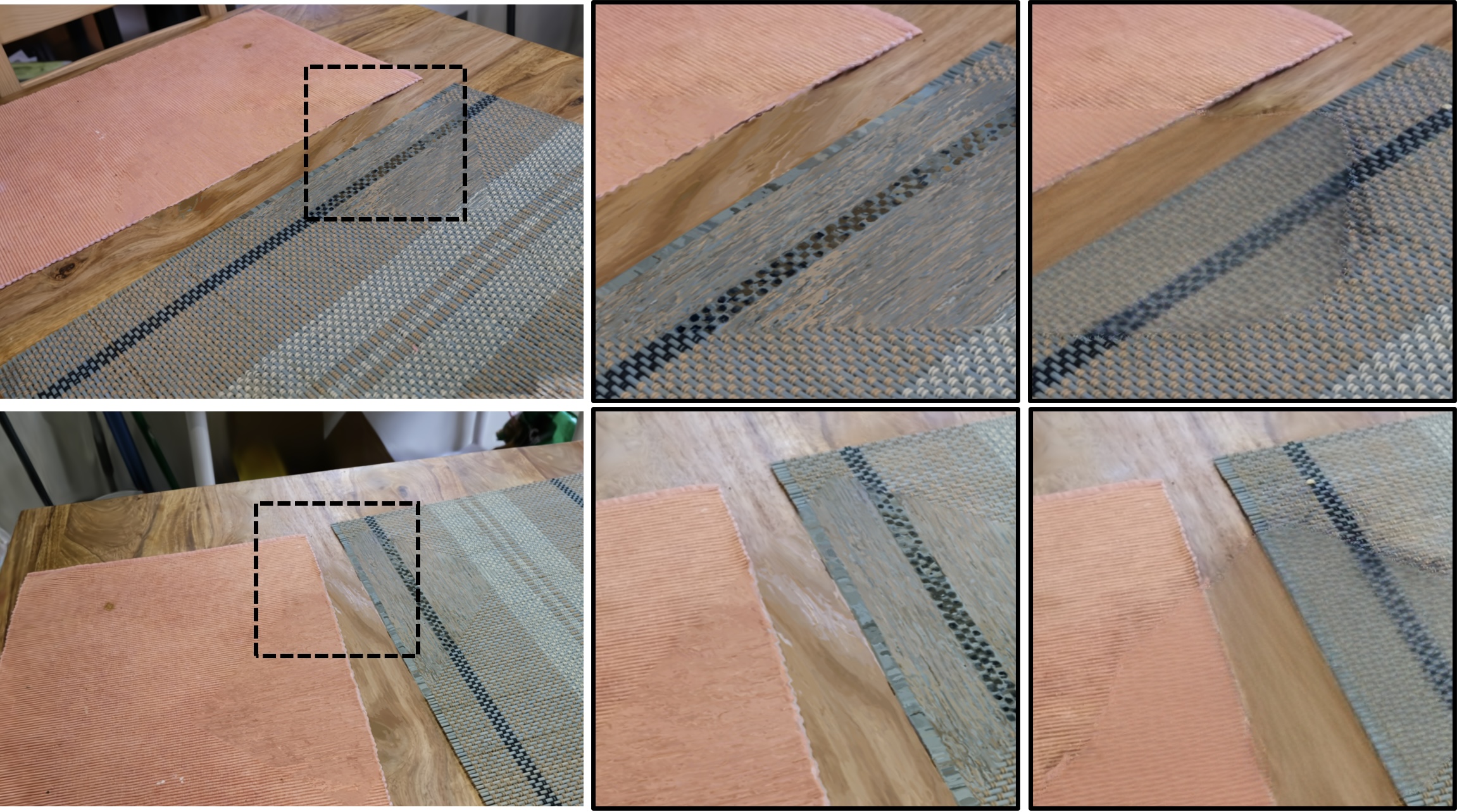}
  \caption{Our method shows a smoother transition when recovering the reflection on the table surface.}
  \label{fig:visual_depent}
\end{figure}

In Figure \ref{fig:visual_depent}, we show the inpainting results of our method alongside AuraFusion360 \cite{wu2025aurafusion} on the Mip-NeRF 360 ``kitchen'' scene. Our method not only achieves a more seamless blending of the ROI with its surroundings, but also significantly improves the reconstruction of specular reflections on the tabletop. By successfully creating a mapping field with consistent Lambertian reflectance, while accurately capturing view-dependent effects, our method outperforms in visual coherence across different viewpoints.

\begin{figure}[t]
  \centering
  \setlength{\tabcolsep}{2pt}
  \renewcommand{\arraystretch}{1.0}
  \begin{tabular}{c ccc}  
    \small Original scene (RGB) & \small AuraFusion360
 & \small Ours \\
    \includegraphics[width=0.32\linewidth, valign=m]{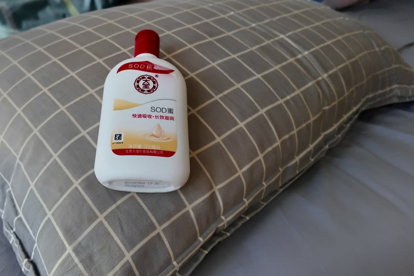} &
    \includegraphics[width=0.32\linewidth, valign=m]{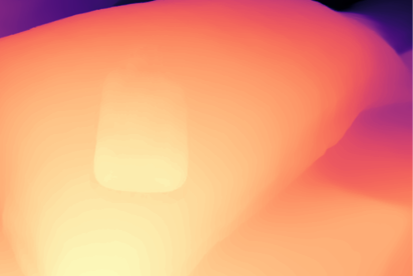} &
    \includegraphics[width=0.32\linewidth, valign=m]{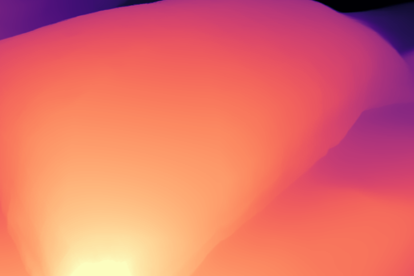}\\[24pt]
    \includegraphics[width=0.32\linewidth, valign=m]{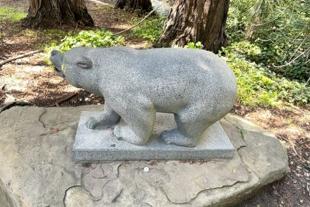} &
    \includegraphics[width=0.32\linewidth, valign=m]{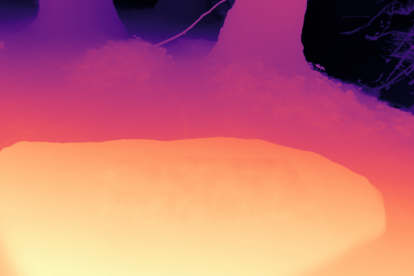} &
    \includegraphics[width=0.32\linewidth, valign=m]{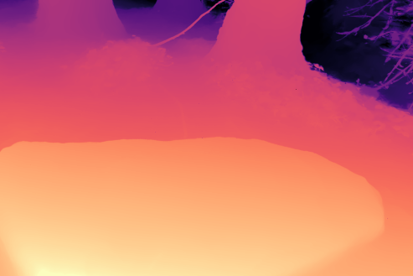} \\
\end{tabular}
\caption{Comparison between depth completion of AuraFusion360 and ours.}
\label{fig:depth_completion}
\end{figure}

We further evaluate our depth completion performance in Figure~\ref{fig:depth_completion}. As demonstrated, our method successfully aligns the incomplete depth with the scale-ambiguous results that are typical for monocular depth estimation models. Our approach effectively handles curved surfaces, maintaining geometric consistency---a critical factor for the subsequent warping steps. By preserving these structural details, we minimize artifacts that often arise from inaccurate depth-to-scene alignment.

\paragraph{Inpainting performance on different resolution scales}
To evaluate the trade-off between quality and efficiency, we perform a quantitative analysis across multiple resolution scales in Figure \ref{fig:resolution_comparison}. Testing is conducted on an NVIDIA RTX 5090 GPU using the IMFine dataset ($1900 \times 1050$). Due to the memory limitations of existing methods, evaluations are capped at $1\times$ scale. Although our processing time increases significantly at higher resolutions—primarily due to the computational overhead of the final Poisson blending—our approach maintains superior PSNR and LPIPS at all scales while remaining more efficient than competing frameworks.

\begin{table}
  \caption{Comparisons between baseline methods and our method. The top table shows results on the IMFine dataset, and the bottom table shows results using AuraFusion360 on the 360‑USID dataset that they proposed. A higher red intensity indicates a better result. Datasets marked with a star denote results copied from the original paper (no code release).}
  \label{table:imfine}
  \centering
  \begin{tabular}{lcccc}
    \toprule
    IMFine & PSNR~($\uparrow$) & LPIPS~($\downarrow$) & FID~($\downarrow$) & Time~($\downarrow$) \\
    \midrule
    SPInNeRF        & 13.75 & 0.456 & 206.43 & 3.0h \\
    NeRFiller        & 17.45 & 0.411 & 71.21 & 53 mins \\
    GScream          & 12.22 & 0.677 & 237.08 & 21 mins \\
    InFusion        & 13.19 & 0.481 & 175.52 & \cellcolor{best} 1 min \\
    IMFine          & \cellcolor{best} *19.67 & \cellcolor{second} *0.268 & *149.52 & *1.0h \\
    AuraFusion360   & 17.46 & \cellcolor{best} 0.173 & \cellcolor{second} 69.33 & 16 mins \\
    Ours (3DGS)     & 17.47 & 0.344 & 87.36 & 10 min \\
    Ours (2DGS)     & \cellcolor{second} 19.09 & \cellcolor{second} 0.248 & \cellcolor{best} 53.11 & \cellcolor{second} 6 mins \\
    \midrule[\heavyrulewidth]
    360-USID & \\
    \midrule
    AuraFusion360          & \cellcolor{best} 16.22 & \cellcolor{second} 0.445 & \cellcolor{second} 61.03 & 32 mins \\
    Ours (3DGS)            & \cellcolor{second} 15.24 & 0.461 & 83.06 & \cellcolor{second} 8.8 mins \\
    Ours (2DGS)            & 15.02 & \cellcolor{best} 0.387 & \cellcolor{best} 57.38 & \cellcolor{best} 6.8 mins \\
    \bottomrule
  \end{tabular}
\end{table}

\subsection{Component Analysis and Extensibility}
\paragraph*{Impact of scaling the guidance field for depth completion}
\label{depth_scale}

While Poisson solvers are effective for depth completion, they often produce over-smoothed results, particularly in irregular regions along Dirichlet boundaries. To address this, we introduce a scaling factor $\mathbf{s_f}$ that adaptively enhances or compresses the gradients derived from the inferred monocular depth map. We evaluate the impact of this factor using the 'rocks' scene from the IMFine dataset \cite{imfine}. Please refer to the figure in the supplemental material. Our observations indicate that the Poisson solver is capable of reconstructing high-fidelity underlying geometry even when faced with complex structural details.
\paragraph*{Impact of Mapping Field on View-Dependent Appearance}

\begin{figure}[t]
  \centering
  \label{pm+diffusion}
  \setlength{\tabcolsep}{2pt}
  \renewcommand{\arraystretch}{1.0}
  \begin{tabular}{c ccc}  
    \small Ours & \small LeftRefill & \small Ours + LeftRefill \\
    \includegraphics[width=0.32\linewidth, valign=m]{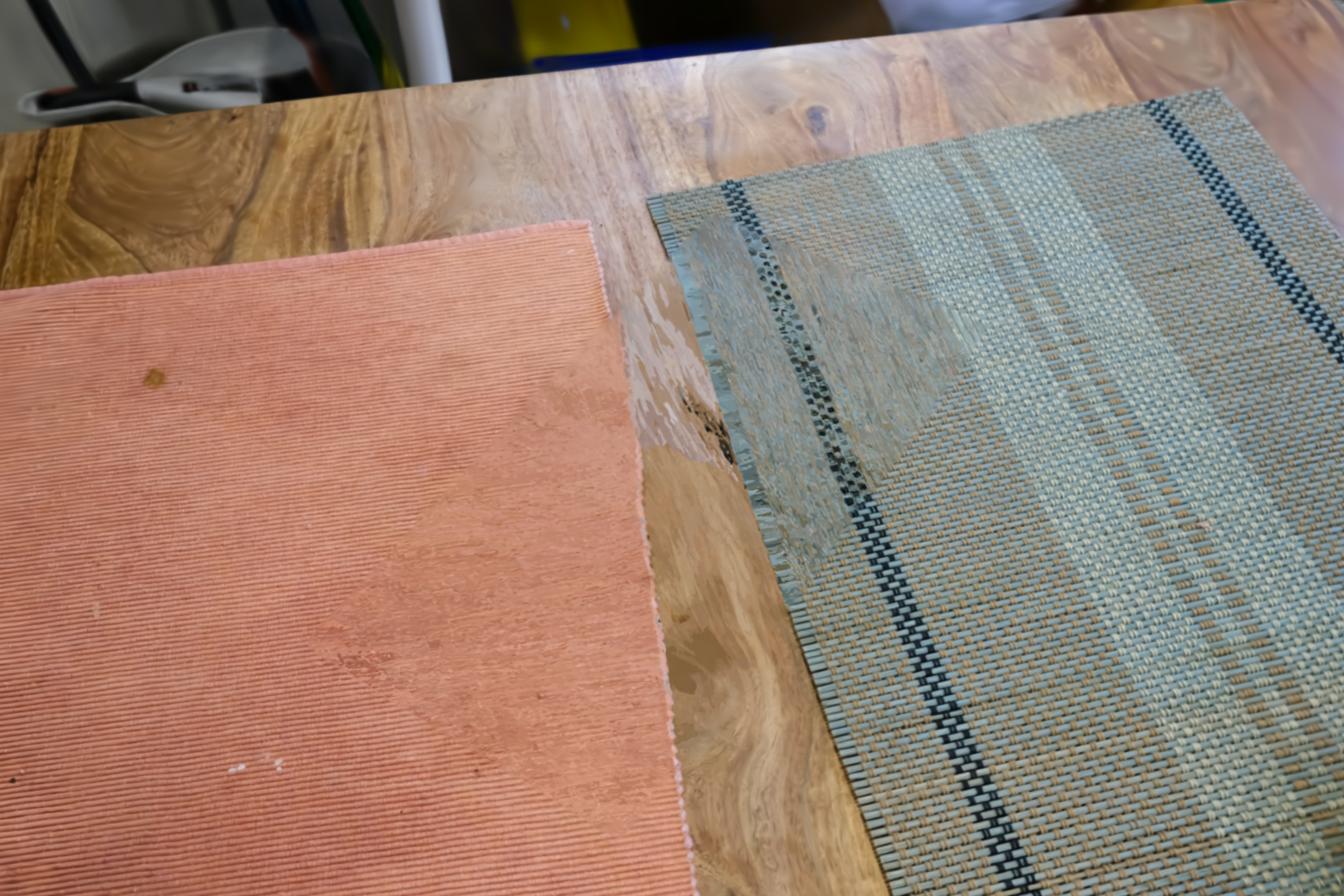} &
    \includegraphics[width=0.32\linewidth, valign=m]{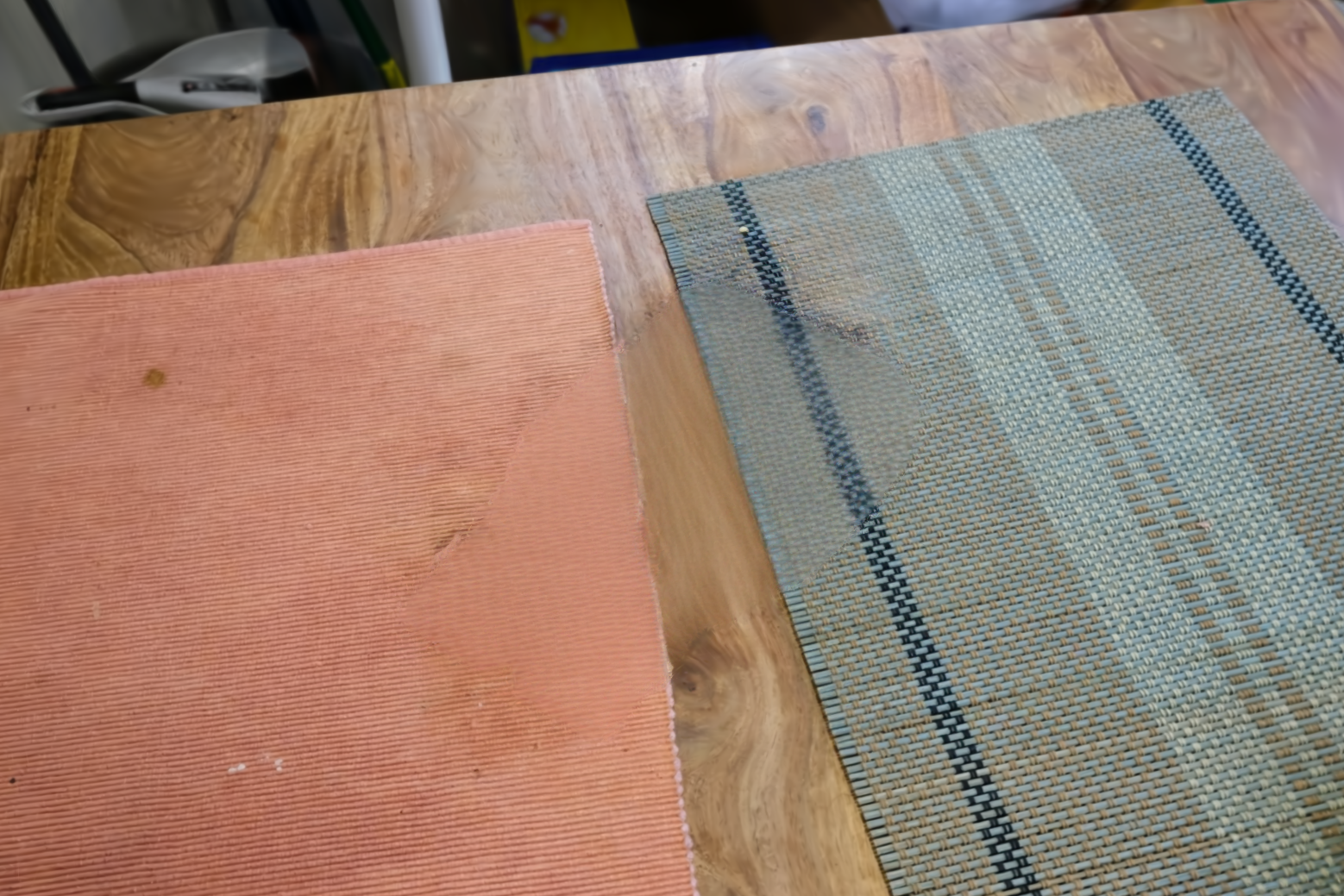} &
    \includegraphics[width=0.32\linewidth, valign=m]{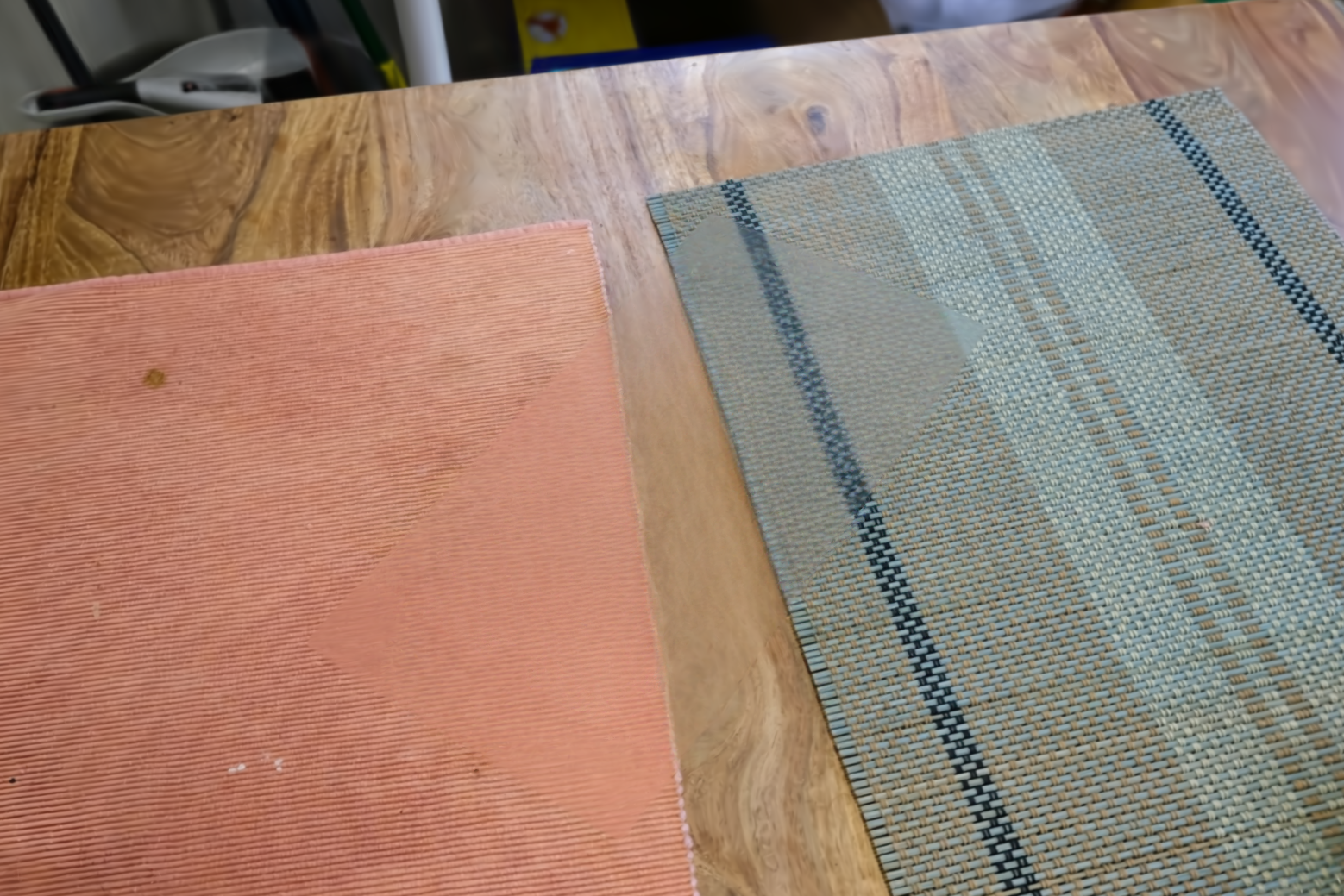}\\[24pt]
    \includegraphics[width=0.32\linewidth, valign=m]{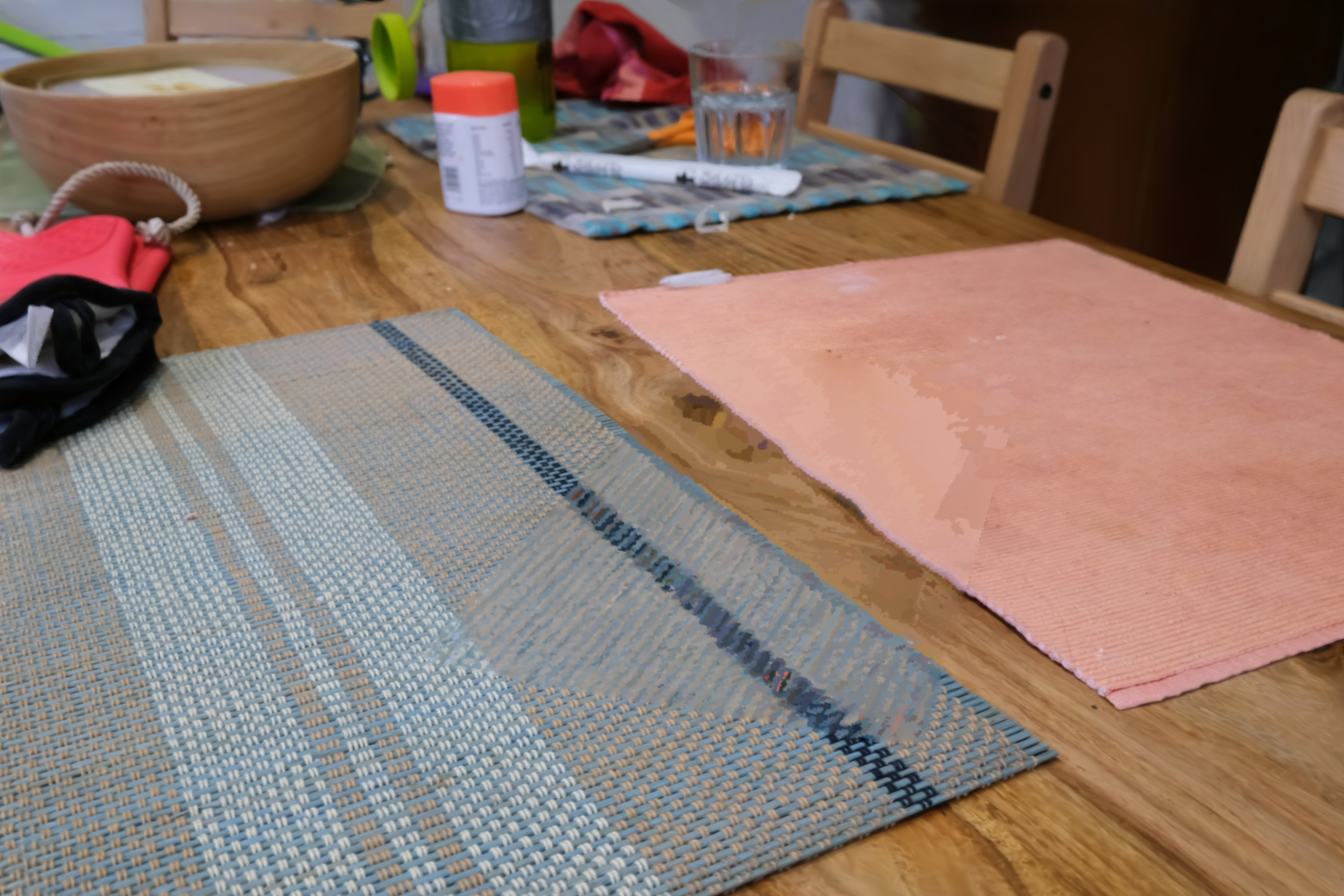} &
    \includegraphics[width=0.32\linewidth, valign=m]{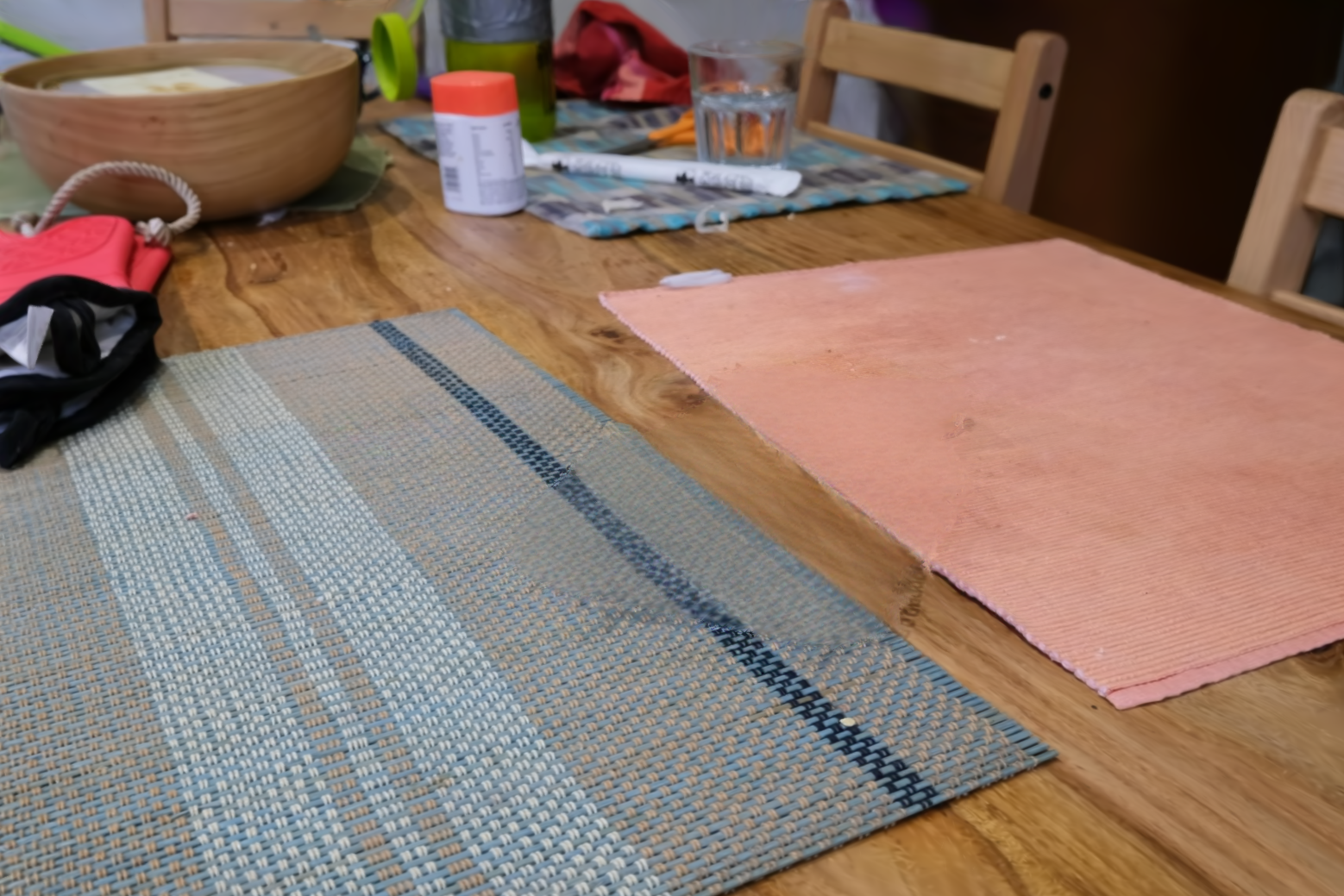} &
    \includegraphics[width=0.32\linewidth, valign=m]{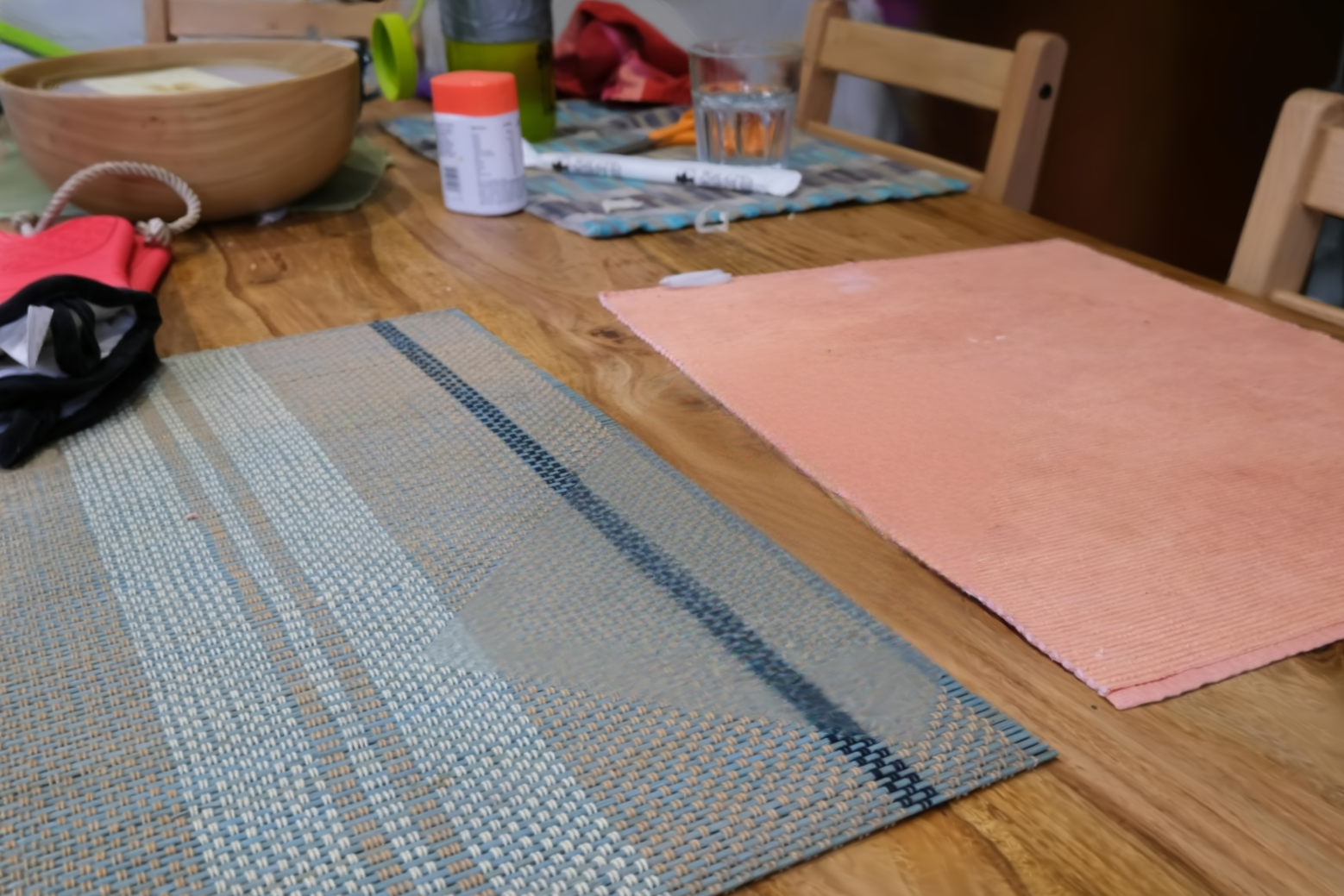} \\
\end{tabular}
\caption{Generative refinement. Our method maintains structural consistency on challenging reflections but may produce artifacts when reference views are limited. While purely diffusion-based methods provide rich texture priors, they suffer from ``averaging'' effects that blur fine details. The hybrid approach leverages structural guidance to mitigate hallucinations, resulting in smoother, higher-fidelity inpainting with fewer artifacts.}
\label{fig:pm_diffusion}
\end{figure}

\begin{table}
  \caption{Structural vs. generative inpainting. Our hybrid framework (Ours + LeftRefill) leverages the geometric foundation of our mapping field to guide generative synthesis. This combination achieves superior perceptual quality (LPIPS) and reconstruction accuracy (PSNR) compared to standalone diffusion baselines, while maintaining competitive view consistency.}
  \label{table:pm+diffusion}
  \centering
  \begin{tabular}{lccc}
    \toprule
    Method & PSNR~($\uparrow$) & LPIPS~($\downarrow$) & FID~($\downarrow$) \\
    \midrule
    LeftRefill~\cite{cao2024leftrefill}          & 16.22 & 0.445 & 61.03\\
    Ours            & 19.09 & 0.248 & \cellcolor{second}53.11 \\
    Ours + LeftRefill   & \cellcolor{second}19.17 & \cellcolor{second}0.227 &  69.99 \\
    \bottomrule
  \end{tabular}
\end{table}

To further validate the effectiveness of our inpainting method, we design ablation studies on the IMFine dataset. \tablename~\ref{table:360_ablation} provides the summary, and the qualitative comparisons can be found in the supplemental materials. In the baseline configuration, we finetune pre-optimized Gaussian primitives with a hole using only a single reference view while masking the ROI in all other views to prevent multi-view supervision. In the second configuration, we warp the reference image to all views with our scale-aligned depth map, but disable sampling from the patch mapping field $\mathbf{f}$. In all cases, the incomplete 2D Gaussian primitives are optimized for 30,000 iterations after novel Gaussian primitives are initialized via depth-based back-projection. 
Our results demonstrate that, while accurate depth-based warping ensures strong view consistency, it fails to recover complex view-dependent effects. By integrating a patch-based mapping sampling strategy, our method successfully inpaints 3D regions while effectively recovering view-dependent appearance that simple warping cannot resolve.

\paragraph*{Patch Matching as a Structural Prior for Generative Refinement}

Beyond internal ablations, we observe that contemporary 3D inpainting frameworks, such as InFusion and AuraFusion360, often introduce structural artifacts due to unaligned Gaussian initialization and inconsistent 2D-to-3D pixel updates. Our Mapping Field (MF) provides a superior structural prior that effectively mitigates these issues.
To verify this claim, we use our inpainted results as source images for a diffusion refinement with the LeftRefill \cite{cao2024leftrefill}. We maintain the exact experimental configuration as in AuraFusion360 for a fair comparison. The result on IMFine dataset and the "ketchen" scene from Mip-NeRF360 is shown in Table \ref{table:pm+diffusion} and \figurename~\ref{fig:pm_diffusion}. We also tested a hybrid configuration which combines our geometric precision with generative high-frequency texture. This \textit{Ours + LeftRefill} configuration significantly reduces hallucinations while maintaining the view consistency established by our mapping field. It achieves the best results (PSNR and LPIPS), albeit at the cost of 3$\times$ or higher runtime.

While the core of our method serves as a highly efficient standalone solution, 
establishes a new state-of-the-art. This hybrid method combines our geometric precision with generative high-frequency texture, significantly reducing hallucinations while maintaining the view consistency.

\subsection{Comparison to Video Diffusion Inpainting Models}

Recent video diffusion models \cite{ren2022look, liu2023zero1to3, shi2023zero123plus, vanhoorick2024gcd} promise superior inpainting by incorporating temporal dimensions and explicit conditioning of camera parameters for 3D video diffusion. These methods also report improved consistency in object removal tasks.

We evaluate our method against GEN3C \cite{ren2025gen3c}, a recent video generation baseline, using the IMFine dataset for object removal. For each scene, we construct the 3D cache to support the rendering of incomplete novel views by randomly selecting five key frames from the training set, including one inpainted frame serving as a ground-truth reference. Our objective is to generate spatially continuous video frames along novel trajectories defined by the test cameras. To ensure a fair comparison, both our approach and the baseline use the same reference frame. The quantitative results are summarized in Table \ref{table:gen3c} and Figures~\ref{fig/gen3c.png}.

\begin{table}
  \caption{Comparison of our method with GEN3C on the IMFine dataset. Red highlights indicate the better performance. To ensure a fair comparison, we only consider the object-mask area, excluding the generated pixels around the ROI, for GEN3C. We show visual results in \figurename~\ref{fig/gen3c.png}.}
  \label{table:gen3c}
  \centering
  \begin{tabular}{lccc}
    \toprule
    Method & PSNR~($\uparrow$) & SSIM($\uparrow$) & LPIPS~($\downarrow$)  \\
    \midrule
    GEN3C~\cite{ren2025gen3c}          & 17.94 & 0.967 & 0.370  \\
    Ours (2DGS)    & \cellcolor{second}19.81 & \cellcolor{second}0.977 & \cellcolor{second}0.238 \\
    \bottomrule
  \end{tabular}
\end{table}

The video diffusion model maintains notably effective multi-view consistency for "dabao". However, the model could not recover similar sharpness in "desk" or "detergent", though reflections are correctly generated in "bin". In contrast, our local patch sampling strategy successfully recreates view-dependent lighting conditions. These complementary strengths suggest that integrating video diffusion with our local patch-based approach can be a promising direction for future work.

\section{Limitations}

While our patch-based 3D inpainting produces high-quality, view-consistent results, our experiments revealed certain limitations. Specifically, while the Poisson blending method is capable of recovering complex surfaces (as discussed in Section \ref{depth_scale}), the optimal scaling factor is currently scene-specific. This suggests that developing a robust, automated scaling mechanism is a promising direction for future research. Furthermore, in extreme cases where the solver fails to reconstruct the underlying geometry, the precision of subsequent patch-warping steps is significantly degraded.

A second limitation concerns appearance-based retargeting after propagating the mapping field. Even with multiple iterations, this method can suffer from weak supervision caused by the lack of ground truth on all target views. Consequently, the method's generalizability is limited in scenes with significant occlusions. A potential solution could directly copy Gaussian primitives into the ROI instead of sampling from a reprojected mapping field. However, this change would require a more structured Gaussian representation, such as Gaussian primitives voxelized by a sparse voxel grid \cite{scaffoldgs, chen2024textto3dusinggaussiansplatting}. Such a structure could provide the necessary spatial priors to maintain 3D consistency where 2D appearance matching becomes ambiguous.

\begin{table}[t]
  \caption{Ablation study on the IMFine dataset. We compare our full method against baselines by adding warping and the patch mapping field individually. Visual comparisons are provided in the supplemental material. 
  }
  \label{table:360_ablation}
  \centering
  \begin{tabular}{cc|ccc}
    \toprule
    Warping & Mapping Field  & PSNR~($\uparrow$) & LPIPS~($\downarrow$) & FID ($\downarrow$) \\
    \midrule
    - & - & 17.48 & 0.571 & 122.47\\
    \checkmark & - &18.54 & 0.251 & 76.21\\ 
    \midrule
    \checkmark & \checkmark & \cellcolor{second}19.09 & \cellcolor{second}0.248 & \cellcolor{second}53.11\\
    \bottomrule
  \end{tabular}
\end{table}
\section{Conclusion and future work}

We present 3D‑GIMP, a novel approach for inpainting 3D Gaussian scenes with the classic, yet highly effective patch‑matching technique. Our approach alleviates the common challenge of multi-view inconsistency typically introduced by the random hallucination of the 2D image generation model, while delivering competitive inpainting quality at significantly reduced processing times. 

Furthermore, our work revisits the task of object removal in 3D Gaussian Splatting, providing a framework for how we could better integrate the prior knowledge from various current foundation models. By bridging these domains, we pave the way for high-fidelity, real-time Diminished Reality~\cite{mori_survey_2017}.

While our method demonstrates robustness in handling consistent lighting and exposure across views, minor artifacts can still emerge in regions with complex view-dependent reflections. Future research will explore geometry-aware patch searching to improve inpainting for complicated lighting or texturing.

\clearpage
\balance
\bibliographystyle{ACM-Reference-Format}
\bibliography{references}

\appendix

\section{Supplementary Material for 3D-GIMP: 3D Gaussian Inpainting Meets PatchMatch}
\subsection{Patch Matching Algorithm}
We provide a pseudo-code block in Algorithm \ref{alg:pyramid_patchmatch} for initiating the patch mapping field, followed by a seminal work, PatchMatch~\cite{patchmatch}.

\begin{algorithm}[ht]
\caption{Multi-scale Patch Mapping Field Construction}
\label{alg:pyramid_patchmatch}
\KwIn{Reference view $I_0$, target region $T_0$, visibility mask $V$, max level $L$, iterations $N$}
\KwOut{Optimized mapping field $\mathbf{f}^{(0)}$}
\BlankLine
$\{I^{(l)},\, V^{(l)}\}_{l=0}^{L} \leftarrow \textsc{Downsample}(I_0,\, V,\, L)$\;
$\mathbf{f}^{(L)} \leftarrow \textsc{RandomInit}\!\left(T_0^{(L)},\, V^{(L)}\right)$\;
\For{$l \leftarrow 0$ \KwTo $L$}{
   \lIf{$l < L$}{$\mathbf{f}^{(L)} \leftarrow \textsc{Upsample}\!\left(\mathbf{f}^{(l+1)}\right)$}
   \For{$\mathit{iter} \leftarrow 1$ \KwTo $N$}{
       \ForEach{$\mathbf{p} \in T_0^{(l)}$}{
           $\mathbf{f}(\mathbf{p}) \leftarrow \textsc{Propagate}\!\left(\mathbf{p},\, \mathbf{f},\, I^{(l)},\, V^{(l)}\right)$\;
           $\mathbf{f}(\mathbf{p}) \leftarrow \textsc{RandomSearch}\!\left(\mathbf{p},\, \mathbf{f}(\mathbf{p}),\, I^{(l)},\, V^{(l)}\right)$\;
       }
   }
}
\Return $\mathbf{f}^{(0)}$\;
\end{algorithm}

\subsection{Initialization of Inpainted Gaussian Primitives}
\label{subsec: gaussian filling}

Per construction, our method achieves strong geometric correlation between multiple viewpoints. Therefore, we can pre-compute SH of all view-dependent colors directly for new Gaussian primitives $G_i$ using constrained weighted least squares. By integrating these with the reference image using the scale-aligned depth map (Section~\ref{subsubsec: depth comple}). We inpaint Gaussian primitives with reduced memory and time requirements during optimization. For brevity, we leave out the primitive index $i$ in the following.

We use the pixel color $c_0$ observed in the reference view $I_0$ to fix the zero-band SH coefficient $h_0=c_0/Y_0$. We want to determine the higher order coefficients from color observations $c_\omega$ in direction $\omega$. Note that we obtain at most one observation per view and per unoccluded point in that view. We assume that the colors observed in other views are similar to the one in $I_0$, so we set $Y_0=0$ to get a residual color
\begin{equation}
r_\omega=c_\omega - c_0 = \sum_{k=0}^K h_k Y_k(\omega) - h_0 Y_0 =\sum_{k=1}^K h_k Y_k(\omega)=\mathbf{h}\mathbf{y}_\omega.
\end{equation}
We stack the residual colors from all observations in a vector of the form $[r_\omega-\mathbf{h}\mathbf{y}_\omega]$. Weighted with a photometric consistency vector $\mathbf{e}$, we obtain a linear system
\begin{equation}
[r_\omega-\mathbf{h}\mathbf{y}_\omega]^\top\text{diag}(\mathbf{e})[r_\omega-\mathbf{h}\mathbf{y}_\omega],
\end{equation}
which we solve for $\mathbf{h}$ in the least squares sense to obtain the SH coefficients needed to initialize new Gaussian primitives. 
The vector $\mathbf{e}=[e_\omega]^\top$ assigns lower weights to samples that exhibit a significant color deviation from the reference prior despite having similar viewing angles. By penalizing these photometric outliers, the solution becomes resilient to noise and artifacts introduced during the local search process in target views. The weights are given as 
\begin{equation}
   e_\omega = \exp \left( -\frac{\| r_\omega \|^2}{2 (\sigma_{0} \cdot \exp(1 - \omega \cdot \omega_0))^2} \right),
\end{equation}
where $\sigma_{0}$ controls the strictness of the weight, and $\omega$ and $\omega_0$ are unit direction vectors of the observations $c_\omega$ and $c_0$, respectively. 

\subsection{Sensitivity test on selection of the reference camera}
\label{sec:camera select}
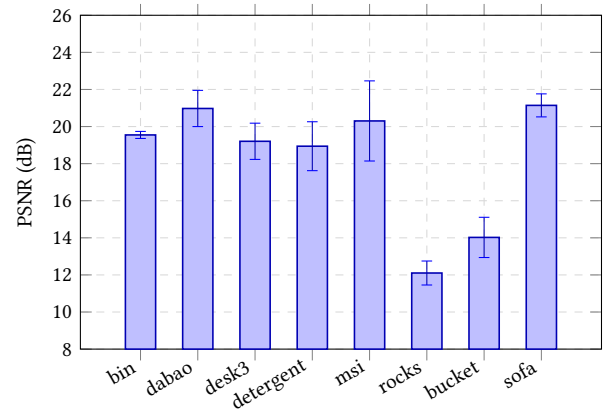
\begin{figure}[ht]
  \centering
  \begin{tikzpicture}
    \begin{axis}[
      ybar,
      width=\columnwidth,
      height=6cm,
      bar width=0.4cm, 
      xtick={1,2,3,4,5,6,7,8},
      xticklabels={bin, dabao, desk3, detergent, msi, rocks, bucket, sofa},
      xticklabel style={font=\small, rotate=30, anchor=east},
      ylabel={PSNR (dB)},
      ylabel style={font=\small},
      yticklabel style={font=\small},
      ymin=8, ymax=26,
      ytick distance=2,
      enlarge x limits=0.15,
      grid=major,
      grid style={dashed, gray!30},
      error bars/y dir=both,
      error bars/y explicit,
    ]
 
    \addplot+[
      fill=blue!25,
      draw=blue!70!black,
      line width=0.6pt,
      error bars/.cd, y dir=both, y explicit,
    ] coordinates {
      (1, 19.5482) +- (0, 0.1896)
      (2, 20.9748) +- (0, 0.9759)
      (3, 19.2052) +- (0, 0.9766)
      (4, 18.9406) +- (0, 1.3187)
      (5, 20.3026) +- (0, 2.1604)
      (6, 12.1033) +- (0, 0.6466)
      (7, 14.0228) +- (0, 1.0844)
      (8, 21.1417) +- (0, 0.6206)
    };
 
    \end{axis}
  \end{tikzpicture}
  \caption{Per-scene PSNR (mean $\pm$ std) across 8 evaluation scenes.}
  \label{fig:psnr_bar}
\end{figure}

\begin{figure}[ht]
  \centering
  \begin{tikzpicture}
    \begin{axis}[
      ybar,
      width=\columnwidth,
      height=6cm,
      bar width=0.4cm,
      xtick={1,2,3,4,5,6,7,8},
      xticklabels={bin, dabao, desk3, detergent, msi, rocks, bucket, sofa},
      xticklabel style={font=\small, rotate=30, anchor=east},
      ylabel={LPIPS Score (Lower is Better)},
      ylabel style={font=\small},
      yticklabel style={font=\small},
      ymin=0, ymax=0.6,
      ytick distance=0.1,
      enlarge x limits=0.15,
      grid=major,
      grid style={dashed, gray!30},
      error bars/y dir=both,
      error bars/y explicit,
    ]
 
 \addplot+[
      fill=red!25, 
      draw=red!70!black,
      line width=0.6pt,
      error bars/.cd, y dir=both, y explicit,
    ] coordinates {
      (1, 0.1223) +- (0, 0.0098)
      (2, 0.2895) +- (0, 0.1300)
      (3, 0.2871) +- (0, 0.0494)
      (4, 0.2584) +- (0, 0.0894)
      (5, 0.2767) +- (0, 0.0358)
      (6, 0.4365) +- (0, 0.0247)
      (7, 0.4773) +- (0, 0.0659)
      (8, 0.3898) +- (0, 0.0139)
    };
 
    \end{axis}
  \end{tikzpicture}
  \caption{Per-scene LPIPS score (mean $\pm$ std) across 8 evaluation scenes.}
  \label{fig:lpips_bar} 
\end{figure}
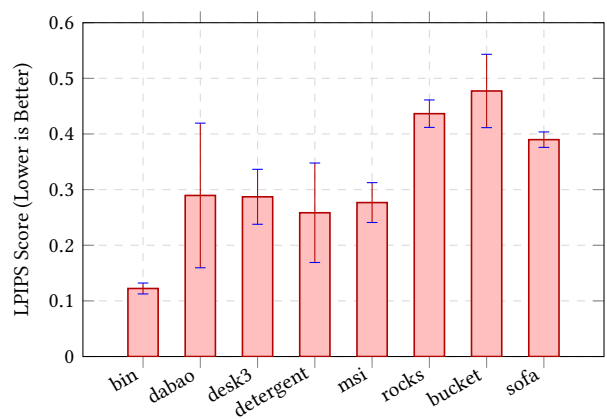

We consistently employ the first camera from the testing set as the reference view to ensure a fair comparison. For in-the-wild datasets, a reference camera is typically selected at random from the available training views. To evaluate the sensitivity of our algorithm to this selection and its impact on the final inpainting quality, we conducted a robust experimental analysis.
Specifically, we performed 10 independent inpainting runs per scene, using different random seeds to vary the reference camera for each viewpoint. We then recorded the mean and standard deviation of the performance metrics across these trials. As illustrated in Figures \ref{fig:psnr_bar} and \ref{fig:lpips_bar}, the choice of reference camera influences inpainting quality. We observed that some scenes are more sensitive to camera choice. 

In the paper, we use the same view as the ground truth for comparison. One possible extension is to choose a reference view $I_0$ whose optical axis is best aligned with the inverse average normal of the removed Gaussian primitives. This viewpoint minimizes perspective foreshortening (See discussions in the literature~\cite{Mori2022GKFtI}). It helps to establish a dense mapping field $\mathbf{f}$ with reduced sampling gaps and stretching artifacts when reprojecting into other views.

\begin{figure*}[t]
  \centering
  \includegraphics[width=\textwidth]{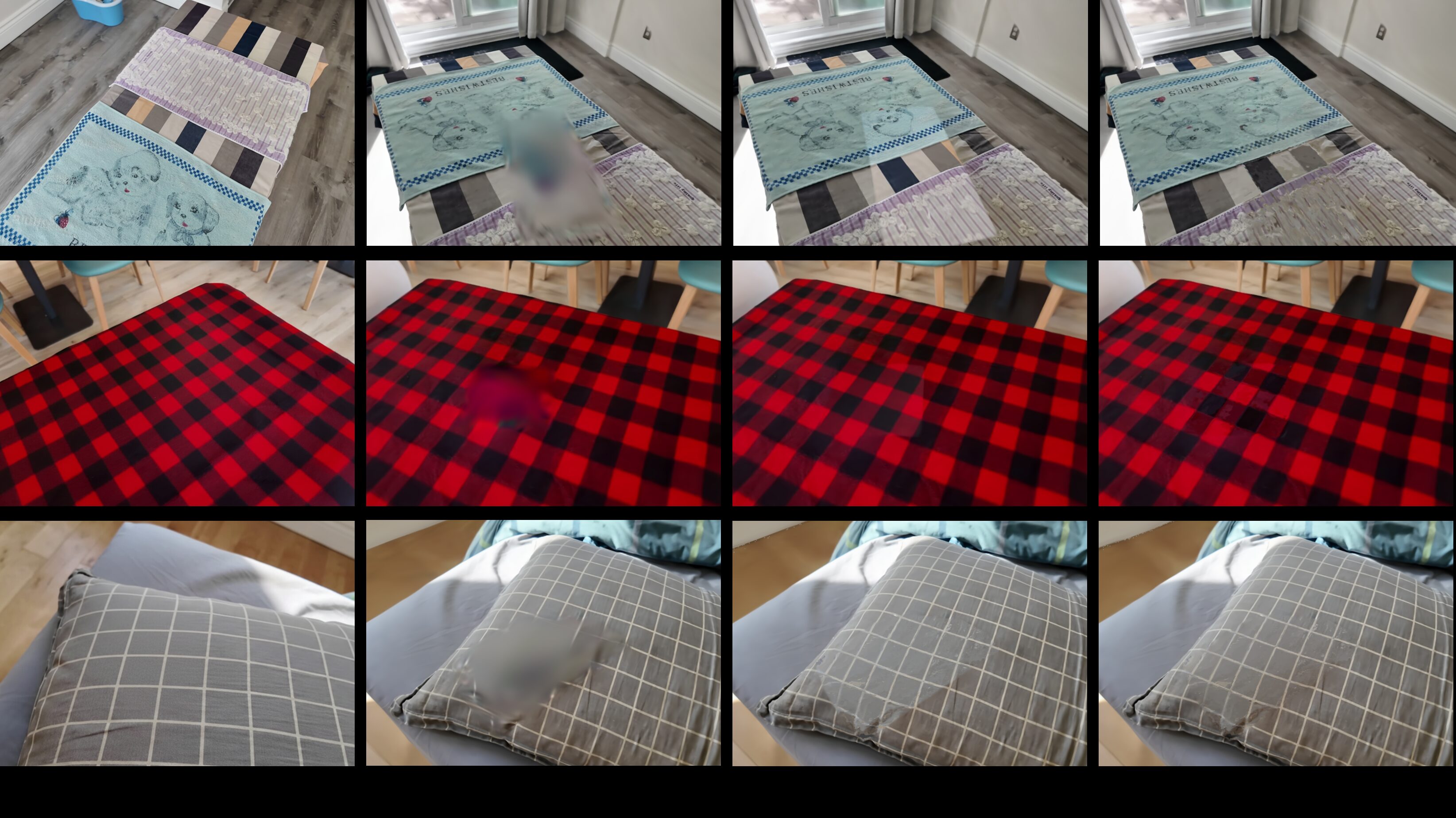}
  \caption{Ablation study results. Our method effectively inpaints the scene by leveraging surrounding pixels from the available views. Even without additional generative models or relighting techniques, our approach produces high‑fidelity, view‑dependent results.
}
  \label{fig:fig/ablation}
\end{figure*}

\begin{figure*}[!t]
  \centering
  \begin{subfigure}{0.18\linewidth}
    \centering
    \includegraphics[width=\linewidth]{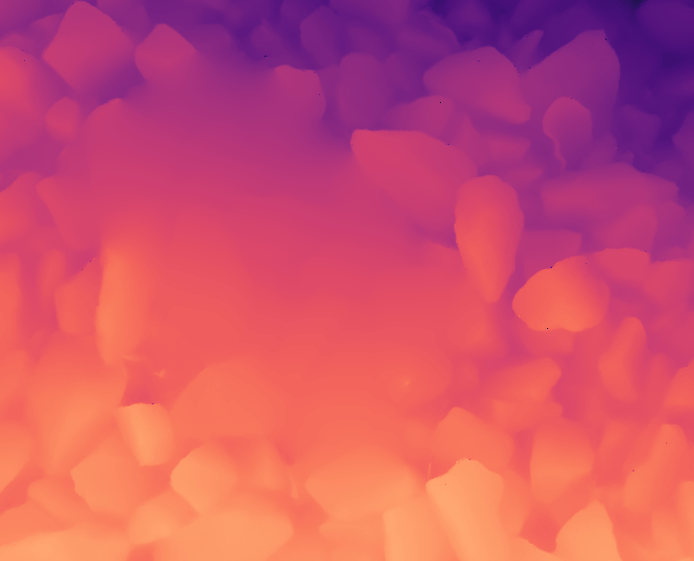}
    \caption{$s_f = 0.01$}
  \end{subfigure}\hfill
  \begin{subfigure}{0.18\linewidth}
    \centering
    \includegraphics[width=\linewidth]{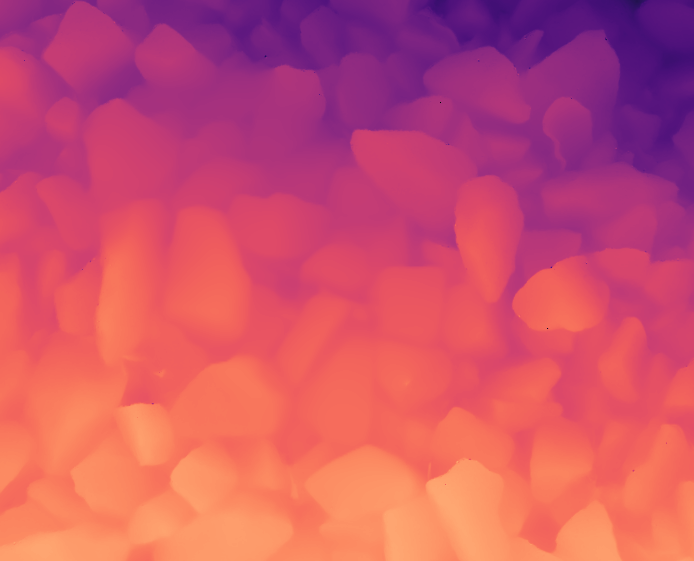}
    \caption{$s_f = 0.05$}
  \end{subfigure}\hfill
  \begin{subfigure}{0.18\linewidth}
    \centering
    \includegraphics[width=\linewidth]{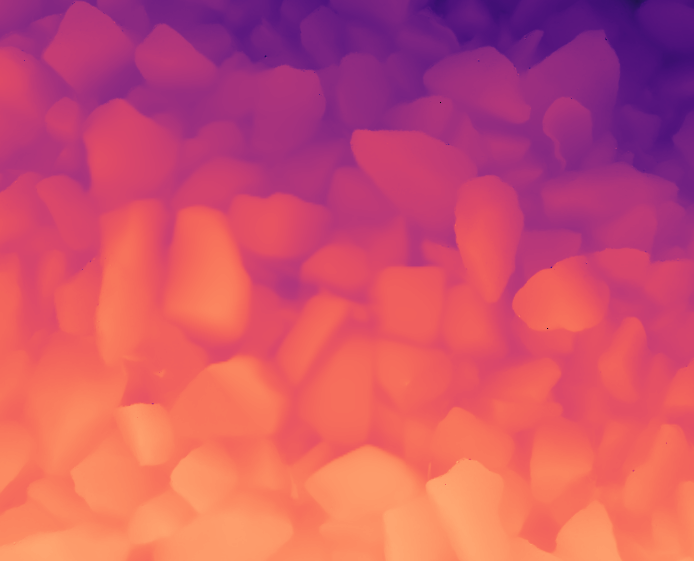}
    \caption{$s_f = 0.10$}
  \end{subfigure}\hfill
  \begin{subfigure}{0.18\linewidth}
    \centering
    \includegraphics[width=\linewidth]{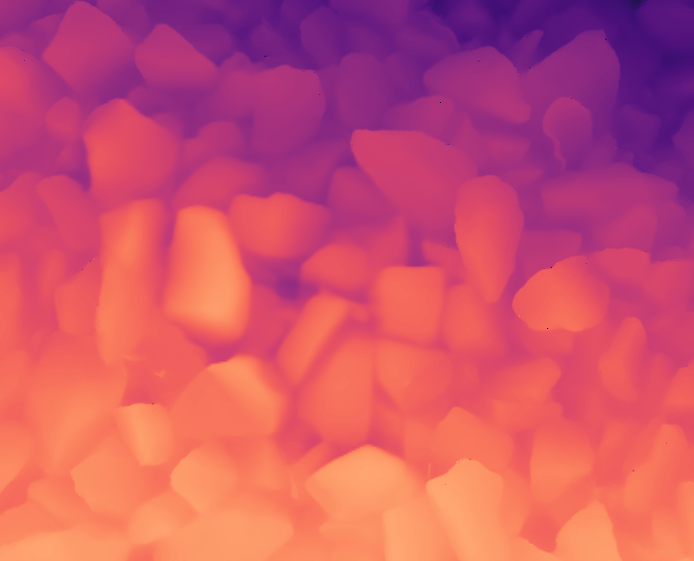}
    \caption{$s_f = 0.20$}
  \end{subfigure}\hfill
  \begin{subfigure}{0.18\linewidth}
    \centering
    \includegraphics[width=\linewidth]{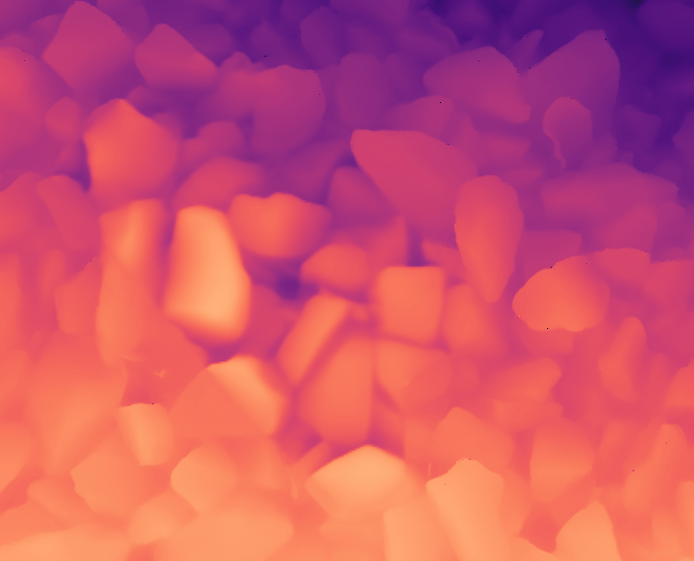}
    \caption{$s_f = 0.50$}
  \end{subfigure}
  \begin{subfigure}{0.045\linewidth}
    \centering
    \begin{tikzpicture}
      \begin{axis}[
        hide axis,
        scale only axis,
        height=2.5cm, 
        width=0.2cm,
        colormap={magma}{
            rgb255=(0,0,5);      
            rgb255=(80,18,123);  
            rgb255=(182,54,121); 
            rgb255=(251,136,97); 
            rgb255=(251,252,191) 
        },
        colorbar,
        colorbar style={
            title={Disp.},
            title style={yshift=-2pt},
            ytick={0,1},
            yticklabels={0.0137, 0.1655},
            tick label style={},
            width=0.2cm,
            yshift=-30pt
        }]
        \addplot [draw=none] {x};
      \end{axis}
    \end{tikzpicture}
    \vspace{8pt} 
  \end{subfigure}
  
  \caption{Impact of the scale parameter $s_f$. Each image shows the disparity map generated by our method for different scale factors. The parameter $s_f$ regulates the influence of the monocular guidance field $d_m$ relative to the boundary constraints. In our experiments, we empirically set $s_f = 0.1$, as this value yielded the most consistent quality across all test scenes.}
  \label{fig:scale_ablation}
\end{figure*}

\subsection{Scale parameter $s_f$ for Depth Completion}
While the Poisson solver is effective for depth completion, they often produce over-smoothed results, particularly in irregular regions along Dirichlet boundaries. To address this, we introduce a scaling factor $\mathbf{s_f}$ that adaptively enhances or compresses the gradients derived from the inferred monocular depth map. We evaluate the impact of this factor using the `rocks' scene from the IMFine dataset~\cite{imfine} (\figurename~\ref{fig:scale_ablation}). Our observations indicate that the Poisson solver can reconstruct high-fidelity underlying geometry even in the presence of complex structural details.

\section{Additional Results}

\begin{figure*}[t]
  \centering
  \includegraphics[width=\textwidth]{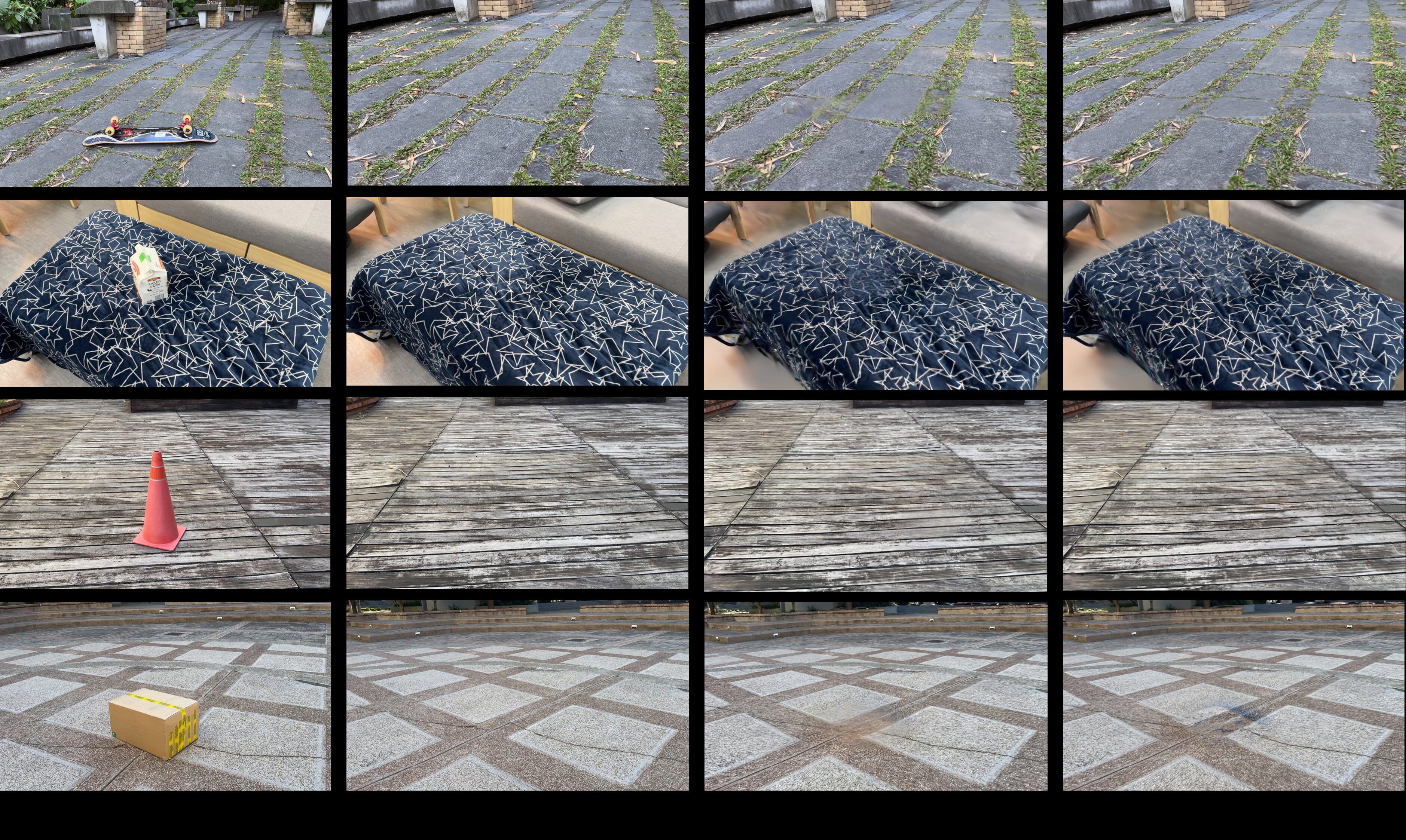}
  \caption{Qualitative results compared with one of most representative 360 inpainting work AuraFusion360}
  \label{fig:fig/360USID_visual}
\end{figure*}

\begin{figure*}[!ht]
  \centering
  \setlength{\tabcolsep}{2pt}
  \renewcommand{\arraystretch}{1.0}
 
\begin{tabular}{c cccc}
 
    &
    \large "bucket" &
    \large "rocks" &
    \large "desk3" &
    \large "sofa" \\
 
    \adjustbox{valign=m}{\rotatebox{90}{\large Original Viewpoint}} &
    \includegraphics[width=0.24\linewidth, valign=m]{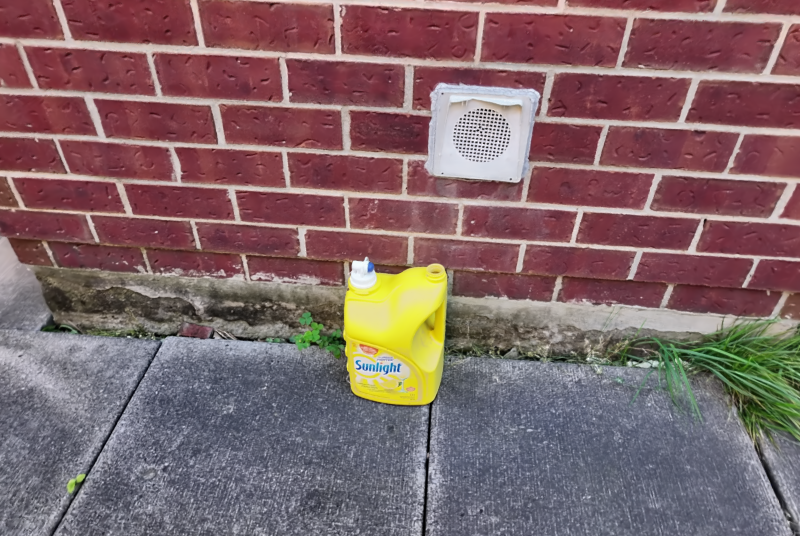} &
    \includegraphics[width=0.24\linewidth, valign=m]{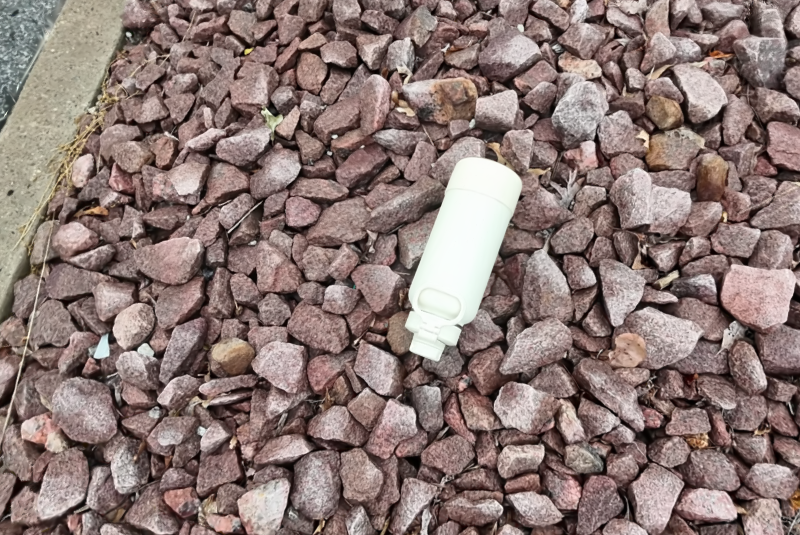} &
    \includegraphics[width=0.24\linewidth, valign=m]{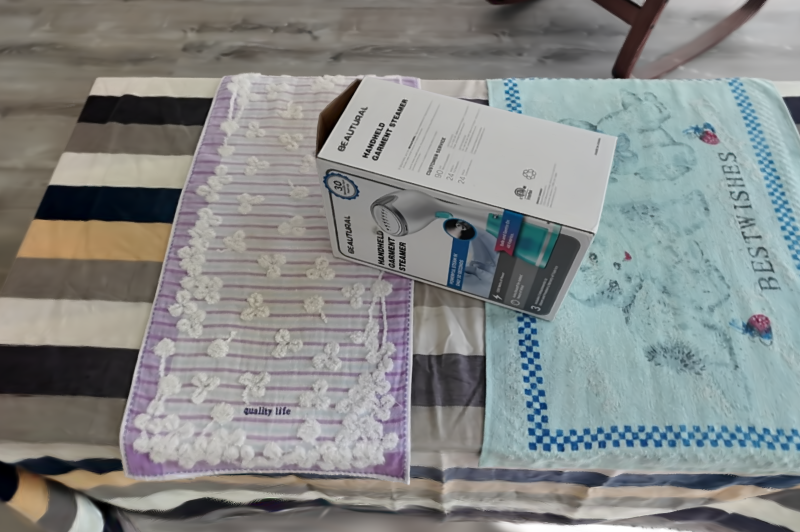} &
    \includegraphics[width=0.24\linewidth, valign=m]{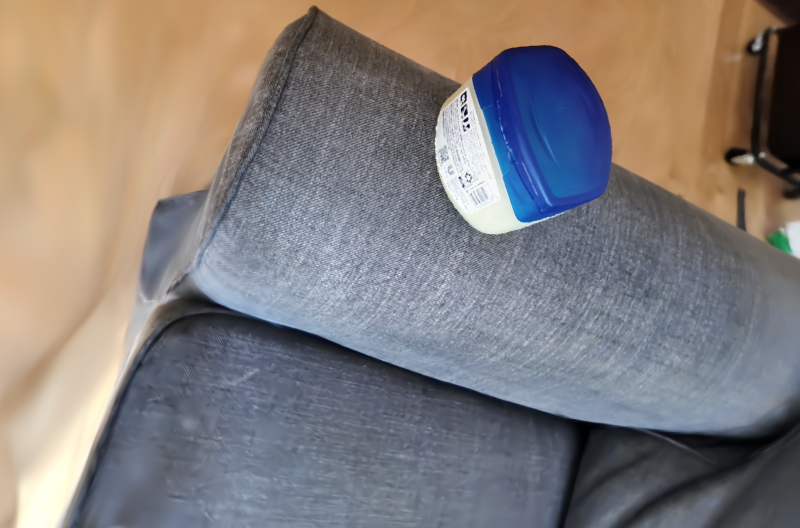} \\[37pt]
    \adjustbox{valign=m}{\rotatebox{90}{\large NeRFiller}} &
    \includegraphics[width=0.24\linewidth, valign=m]{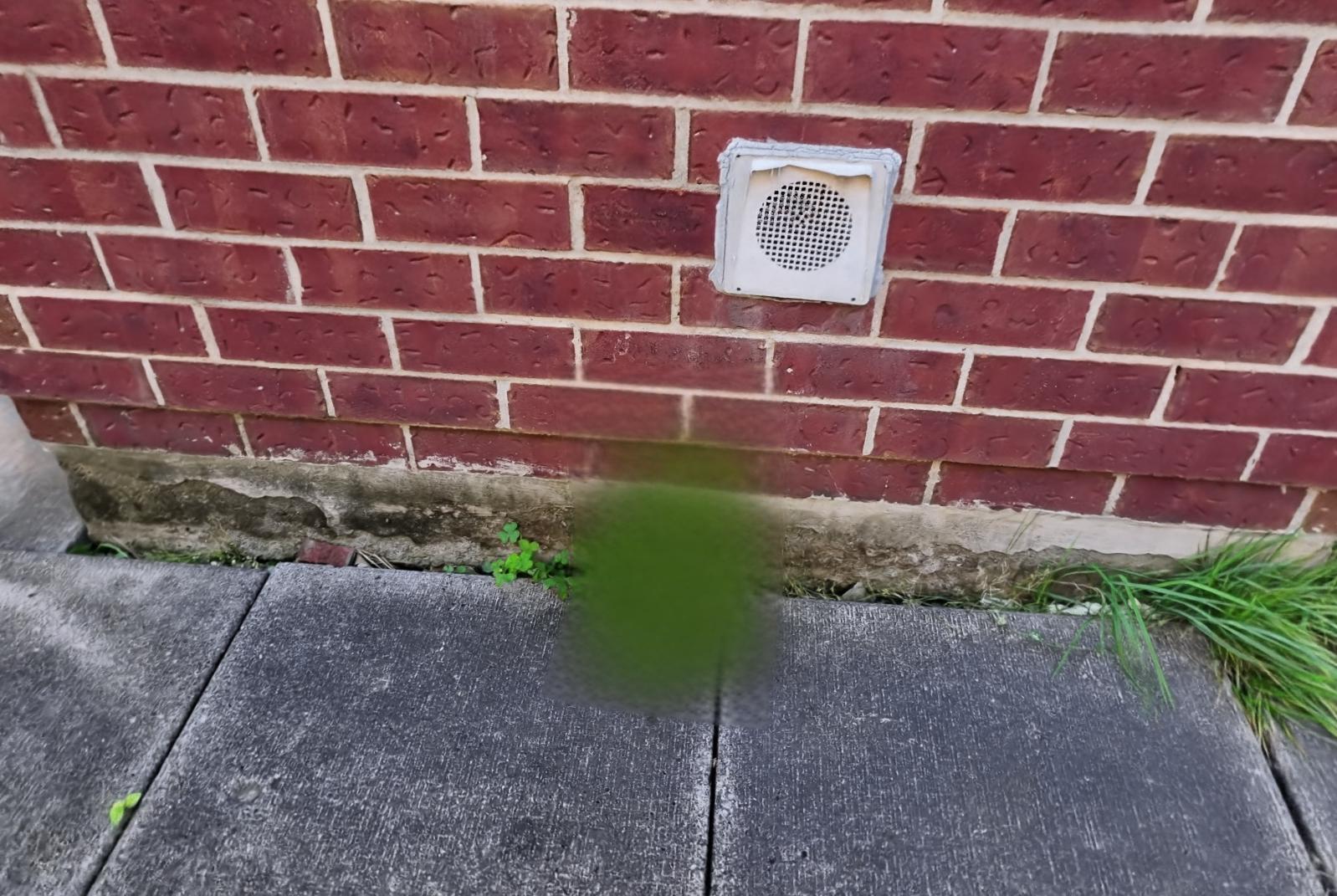} &
    \includegraphics[width=0.24\linewidth, valign=m]{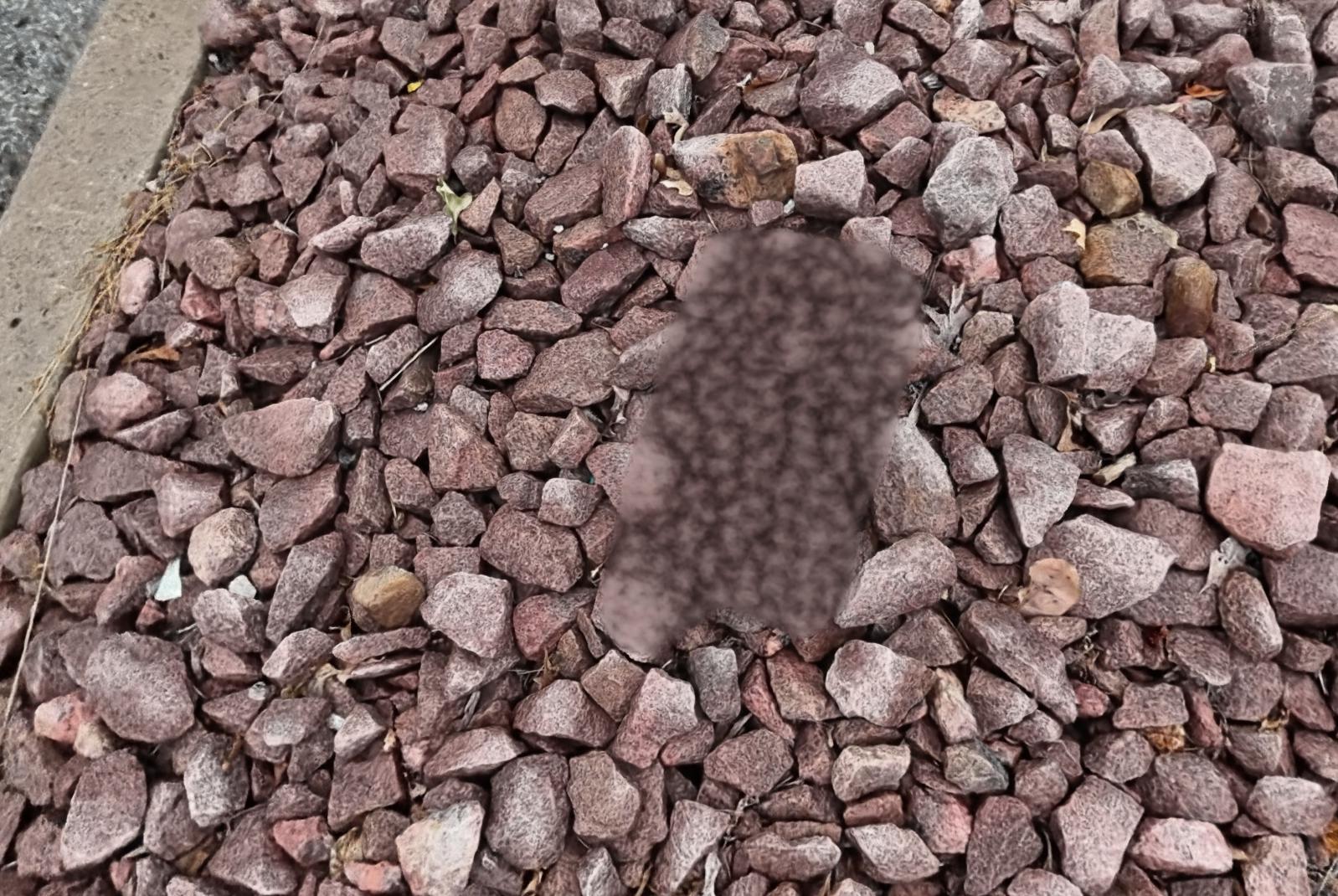} &
    \includegraphics[width=0.24\linewidth, valign=m]{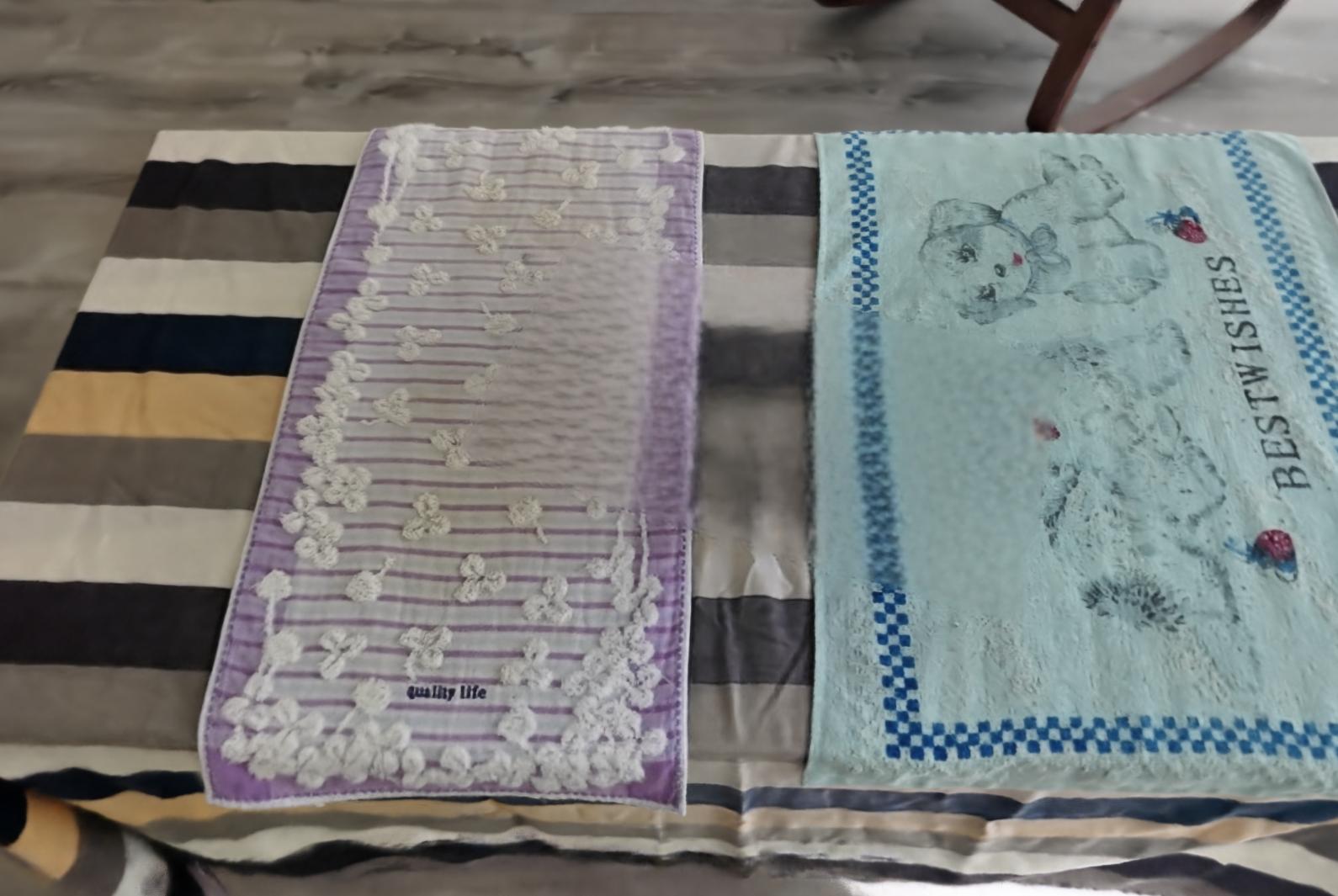} &
    \includegraphics[width=0.24\linewidth, valign=m]{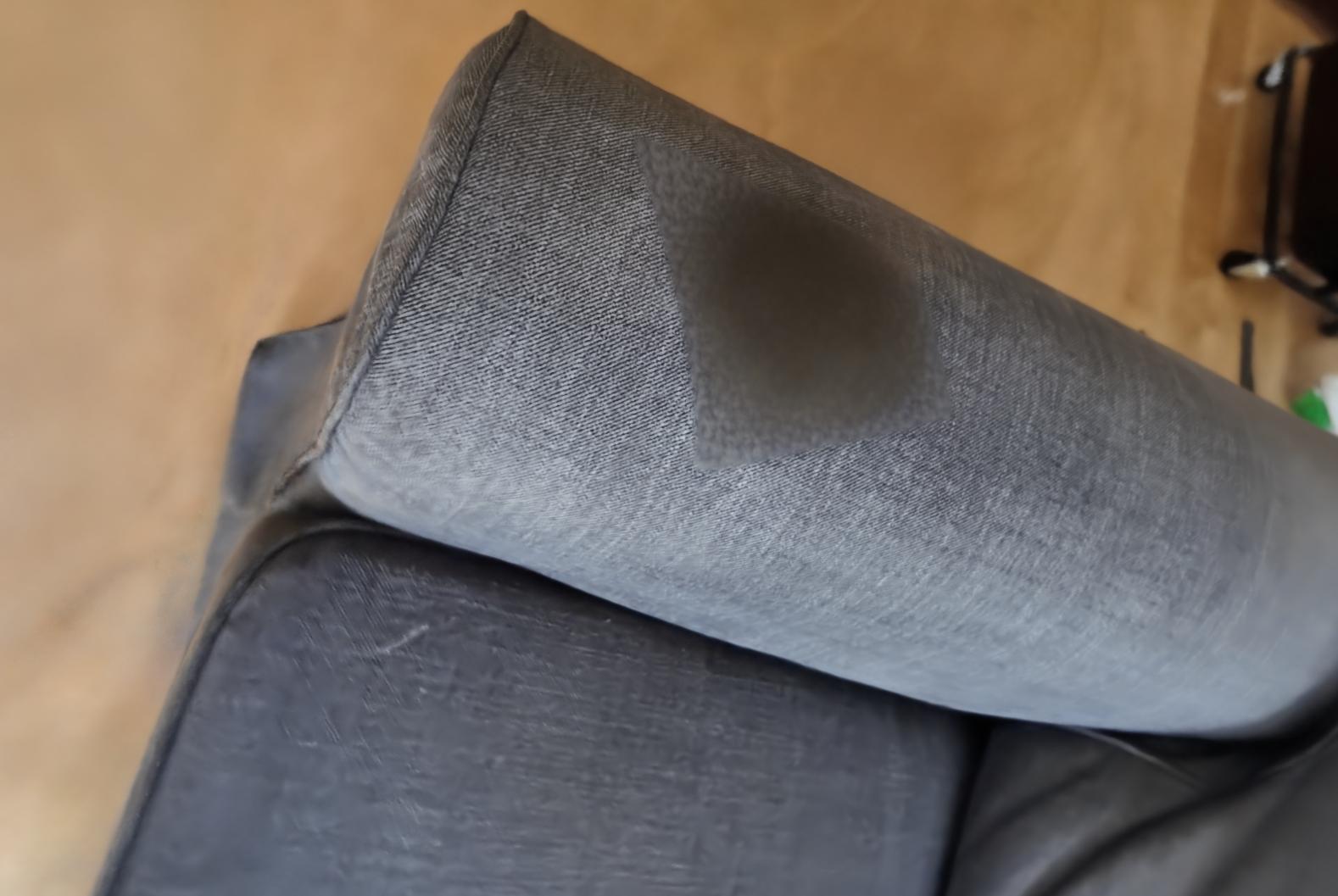} \\[37pt]
    
    \adjustbox{valign=m}{\rotatebox{90}{\large InFusion}} &
    \includegraphics[width=0.24\linewidth, valign=m]{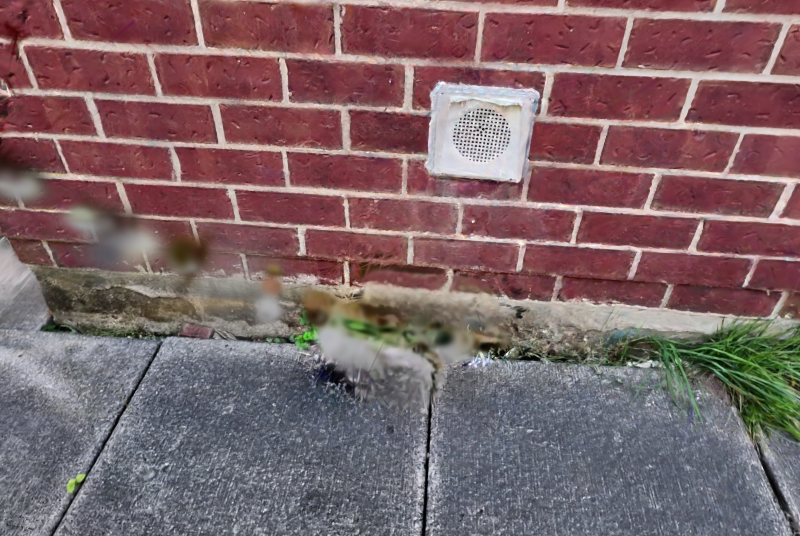} &
    \includegraphics[width=0.24\linewidth, valign=m]{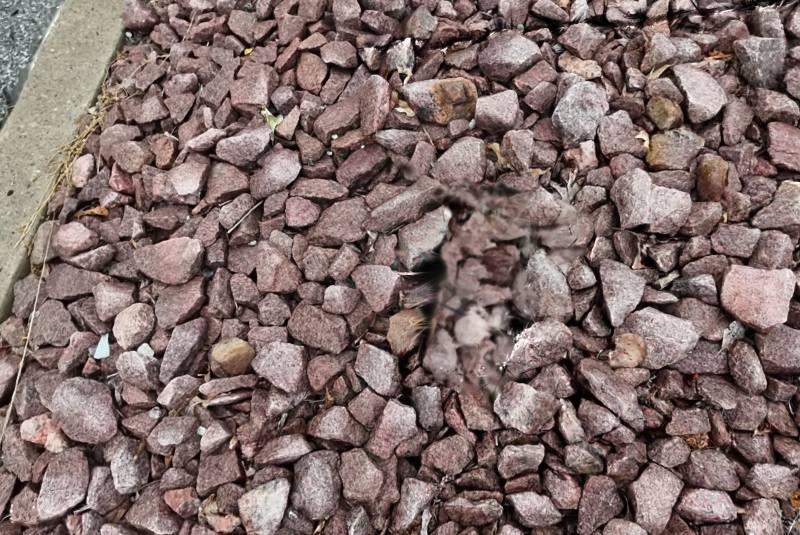} &
    \includegraphics[width=0.24\linewidth, valign=m]{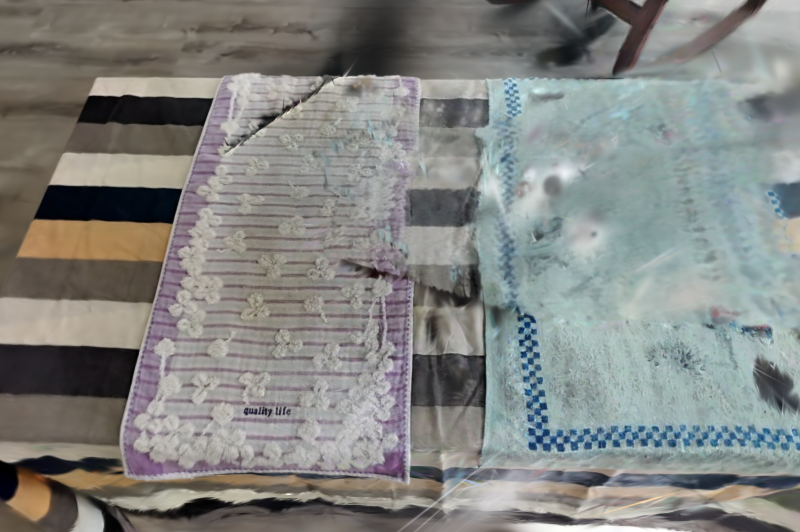} &
    \includegraphics[width=0.24\linewidth, valign=m]{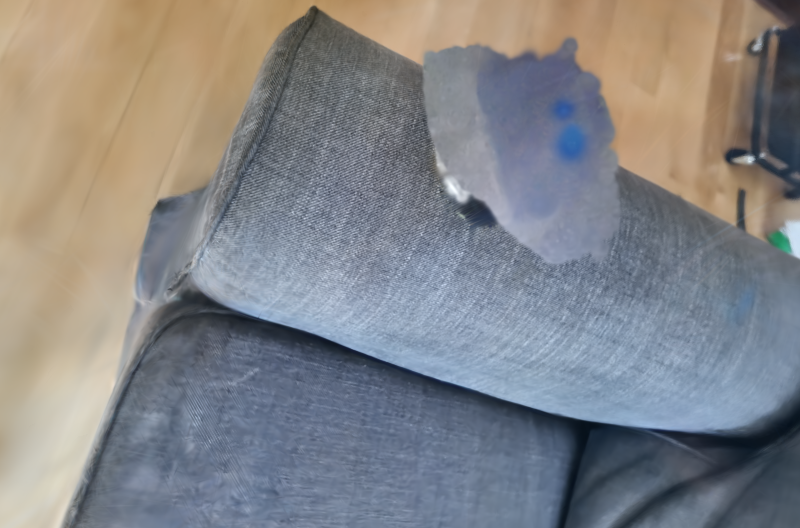} \\[37pt]
 
    \adjustbox{valign=m}{\rotatebox{90}{\large AuraFusion360}} &
    \includegraphics[width=0.24\linewidth, valign=m]{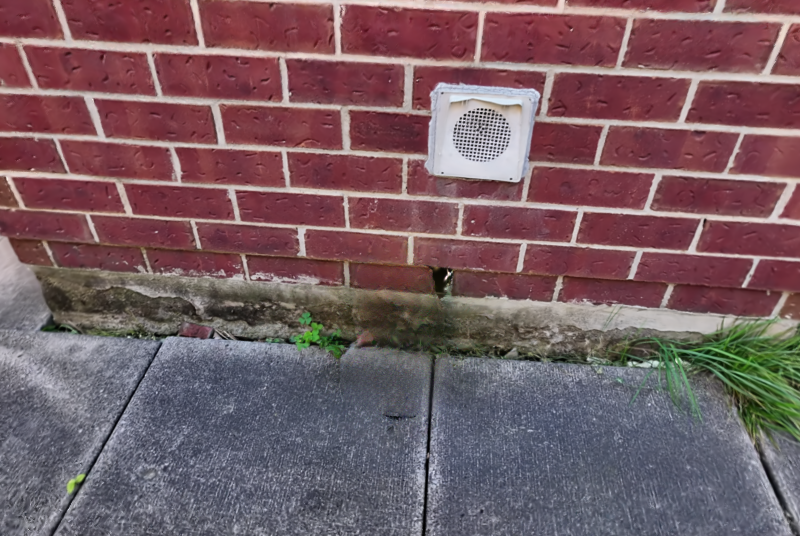} &
    \includegraphics[width=0.24\linewidth, valign=m]{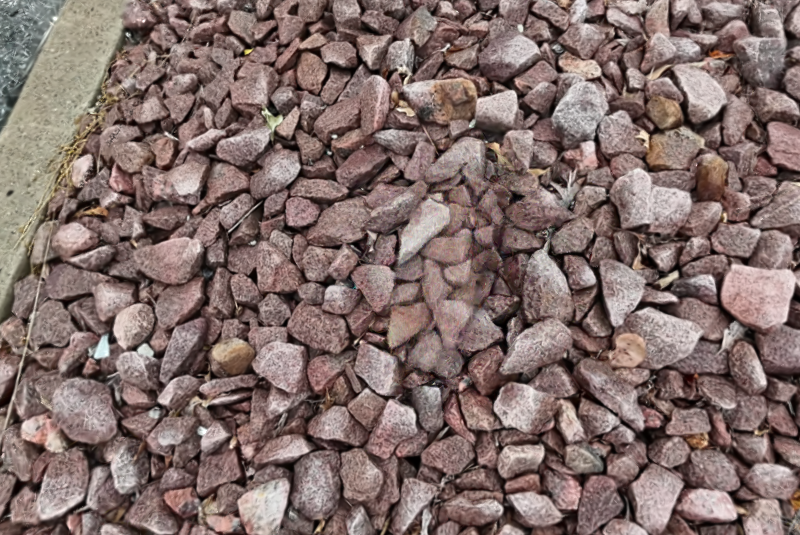} &
    \includegraphics[width=0.24\linewidth, valign=m]{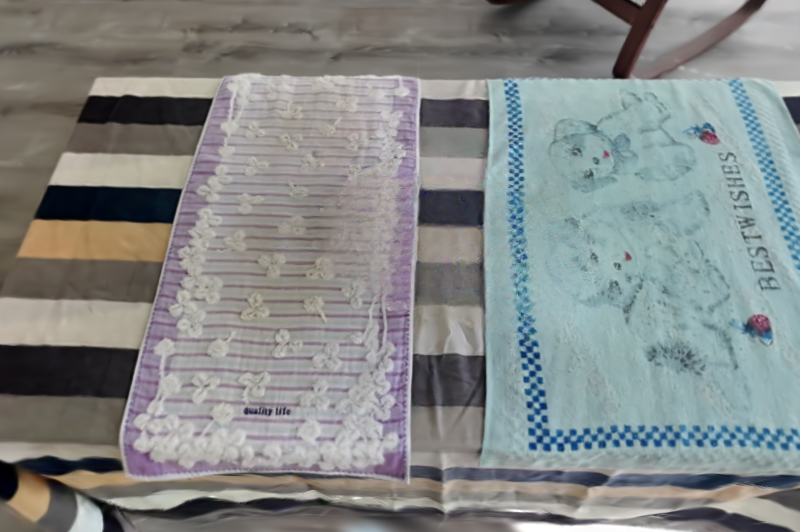} &
    \includegraphics[width=0.24\linewidth, valign=m]{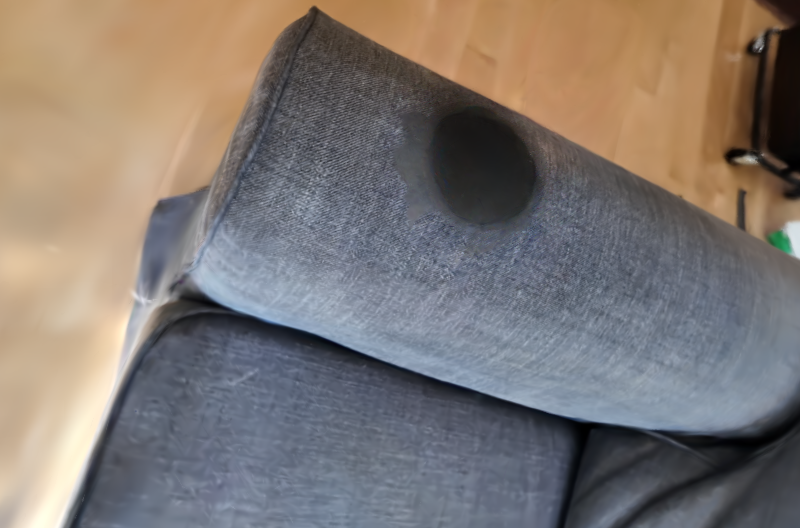} \\[37pt]
 
    \adjustbox{valign=m}{\rotatebox{90}{\large Ours}} &
    \includegraphics[width=0.24\linewidth, valign=m]{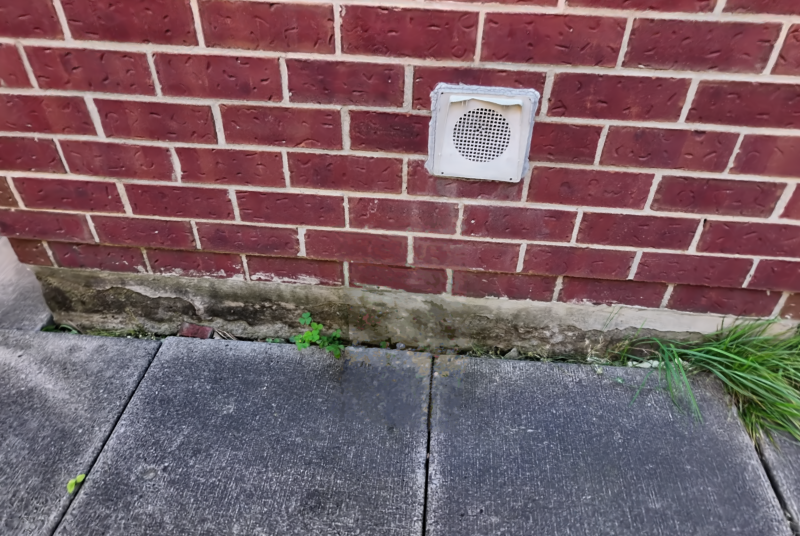} &
    \includegraphics[width=0.24\linewidth, valign=m]{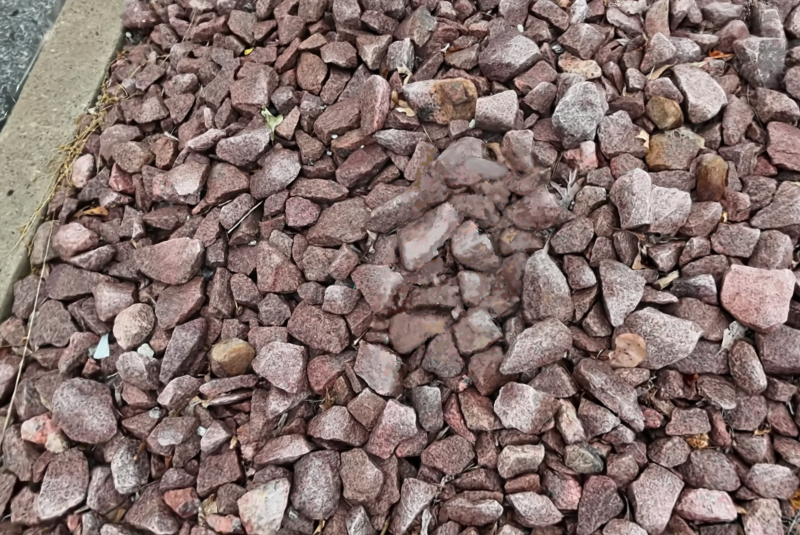} &
    \includegraphics[width=0.24\linewidth, valign=m]{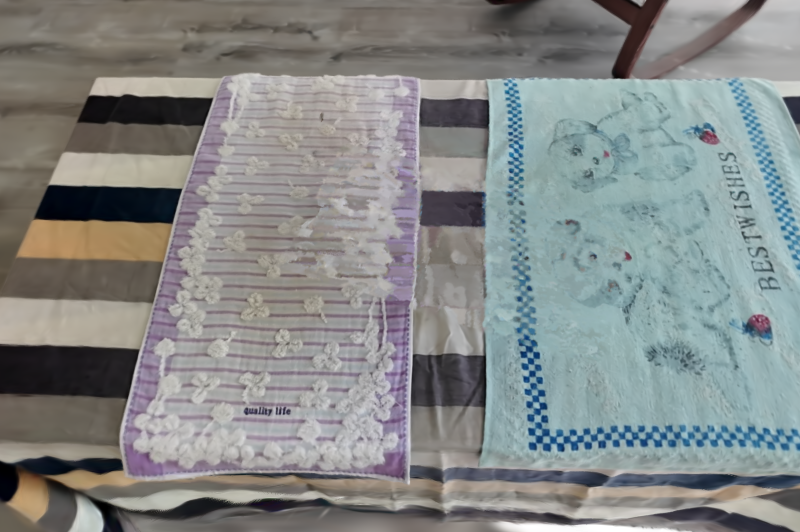} &
    \includegraphics[width=0.24\linewidth, valign=m]{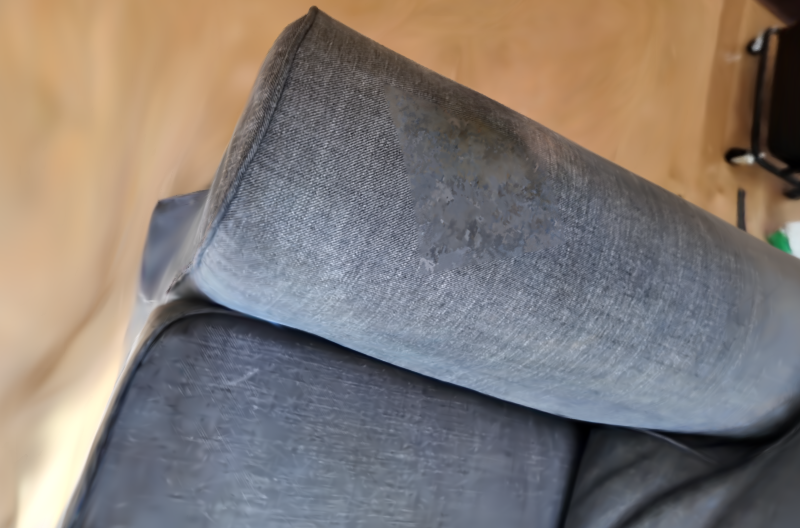} \\[37pt]
 
    \adjustbox{valign=m}{\rotatebox{90}{\large Ground truth}} &
    \includegraphics[width=0.24\linewidth, valign=m]{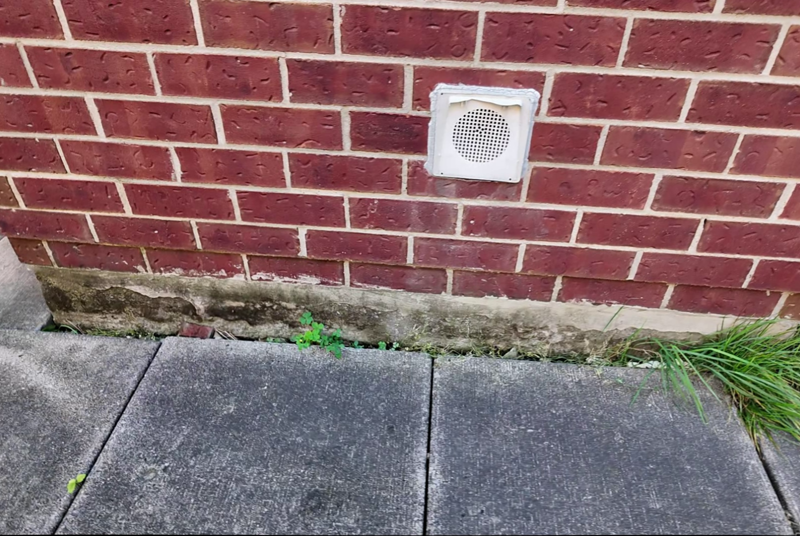} &
    \includegraphics[width=0.24\linewidth, valign=m]{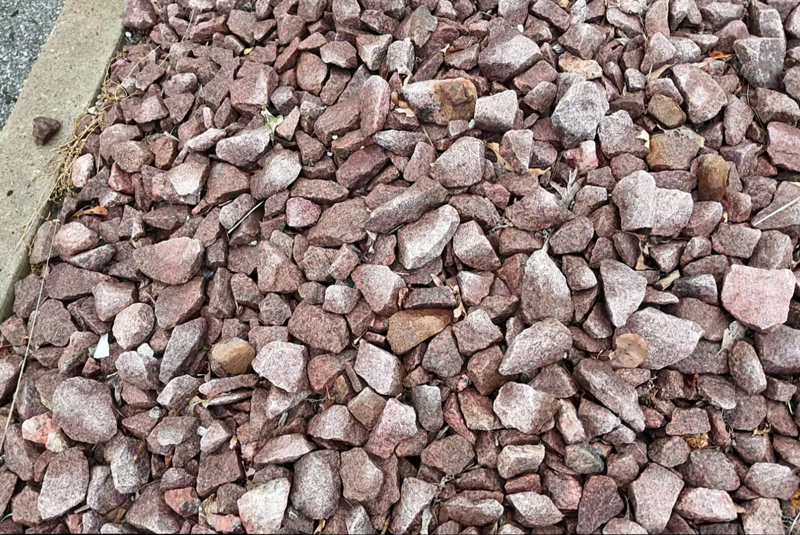} &
    \includegraphics[width=0.24\linewidth, valign=m]{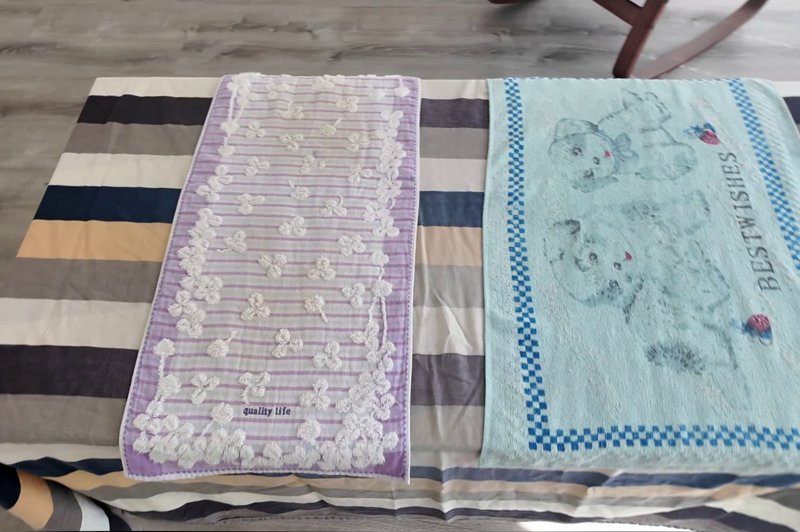} &
    \includegraphics[width=0.24\linewidth, valign=m]{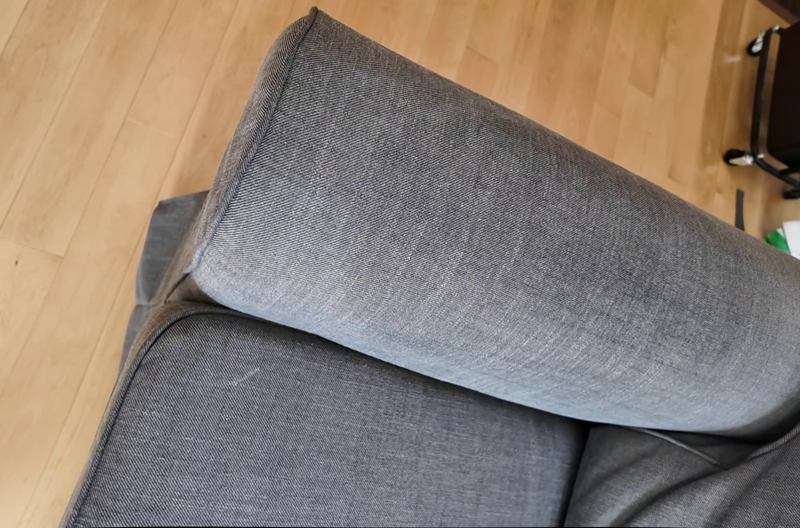} \\
 
\end{tabular}
  \caption{We provide qualitative comparisons against state-of-the-art methods on IMFine dataset. While InFusion \cite{liu2024infusion} achieves the fastest inpainting fine-tuning, its performance is significantly hampered by misaligned depth completion, which leads to visible artifacts. AuraFusion360 \cite{wu2025aurafusion} yields promising initial results on some parts of scenes, but per-frame diffusion inpainting increases the likelihood of detail hallucination, ultimately degrading visual consistency.}
  \label{fig/imfine_render.png}
  
\end{figure*}

 
\begin{figure*}[!ht]
  \centering
  \pgfplotsset{
    resbar/.append style={
      bar width        = 5pt,     
      height           = 5.0cm,
      enlarge x limits = {abs=14pt},
    },
  }
 
  \begin{subfigure}[t]{0.33\linewidth}
    \centering
    \begin{tikzpicture}
      \begin{axis}[
        resbar,
        title        = {(a)~PSNR (dB) $\uparrow$},
        ymin         = 4,  ymax = 24,
        ytick        = {0,6,12,18,24},
        ylabel       = {dB},
        ylabel style = {font=\footnotesize, yshift=-3pt},
        legend style = {
          at            = {(0.5, 1.20)},
          anchor        = south,
          legend columns= 3,
          font          = \scriptsize,
          draw          = gray!40,
          fill          = white,
          inner sep     = 3pt,
          column sep    = 6pt,
        },
      ]
        \plotFourEighty{(0,12.22636)(1,18.19)(2,15.9)(3,20.31)}
        \addlegendentry{$4\times$}
        \plotTenEighty {(0,12.22)(1,17.45)(2,17.46)(3,19.81)}
        \addlegendentry{$2\times$}
        \plotFourK     {(0,13.31366)(1,17.17)(2,15.45)(3,19.47)}
        \addlegendentry{$1\times$}
      \end{axis}
    \end{tikzpicture}
  \end{subfigure}\hfill
  \begin{subfigure}[t]{0.33\linewidth}
    \centering
    \begin{tikzpicture}
      \begin{axis}[
        resbar,
        title        = {(b)~LPIPS $\downarrow$},
        ymin         = 0.0, ymax = 0.85,
        ytick        = {0,0.2,0.4,0.6,0.8},
        ylabel       = {},
        legend style = {draw=none, fill=none},   
      ]
        \plotFourEighty{(0,0.7093)(1,0.2532)(2,0.340325)(3,0.167)}
        \plotTenEighty {(0,0.667)(1,0.411)(2,0.173013)(3,0.238)}
        \plotFourK     {(0,0.631988)(1,0.425)(2,0.407)(3,0.289)}
      \end{axis}
    \end{tikzpicture}
  \end{subfigure}\hfill
  \begin{subfigure}[t]{0.33\linewidth}
    \centering
    \begin{tikzpicture}
      \begin{axis}[
        resbar,
        title        = {(c)~Time (min) $\downarrow$},
        ymin         = 0, ymax = 115,
        ytick        = {0,30,60,90},
        ylabel       = {min},
        ylabel style = {font=\footnotesize, yshift=-3pt},
        legend style = {draw=none, fill=none},   
      ]
        \plotFourEighty{(0,16)(1,29)(2,14)(3,4.22)}
        \plotTenEighty {(0,21)(1,53)(2,16)(3,6)}
        \plotFourK     {(0,30)(1,190)(2,20)(3,21)}
      \end{axis}
    \end{tikzpicture}
  \end{subfigure}
 
  \caption{%
    Comparison of 3DGS inpainting methods across three
    downsampled resolution ($1\times$, $2\times$, $4\times$).
    Each group of bars represents one method; bar color and hatch pattern encode the rendering scale (legend in~(a)). We ensure a fair comparison by recording processing times only from the moment the incomplete rendered image and inpainting mask are created until the 3D Gaussian inpainting is complete, consistent with the methodology in Sec. \ref{sect:4.3}. \textbf{Ours} achieves the best PSNR at all scales while
    requiring substantially less processing time.%
  }
  \label{fig:resolution_comparison}
\end{figure*}
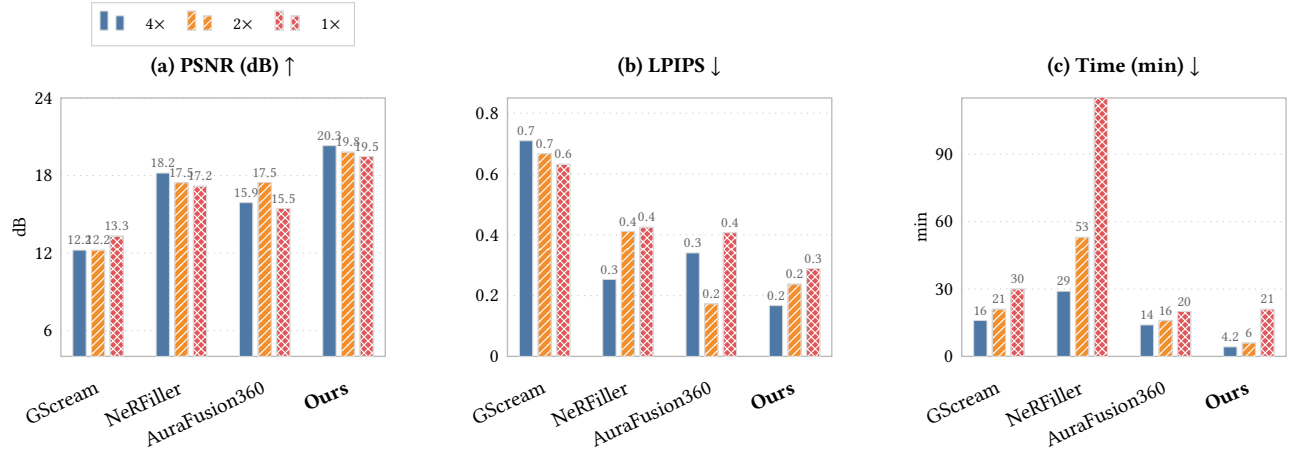

\begin{figure*}[t]
  \centering
  \setlength{\tabcolsep}{2pt}
  \renewcommand{\arraystretch}{1.0}
 
\begin{tabular}{c cccc}
    &
    \large "dabao" &
    \large "detergent" &
    \large "desk3" &
    \large "bin" \\[4pt]
    \adjustbox{valign=m}{\rotatebox{90}{\large GEN3C}} &
    \includegraphics[width=0.24\linewidth, valign=m]{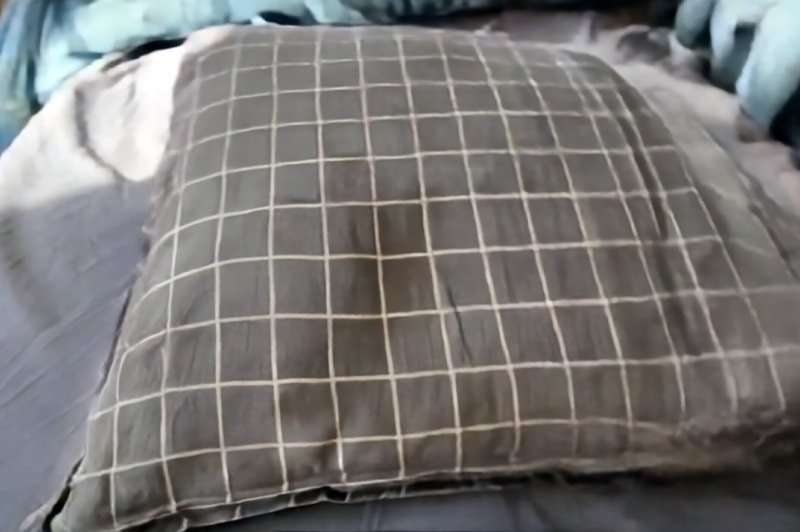} &
    \includegraphics[width=0.24\linewidth, valign=m]{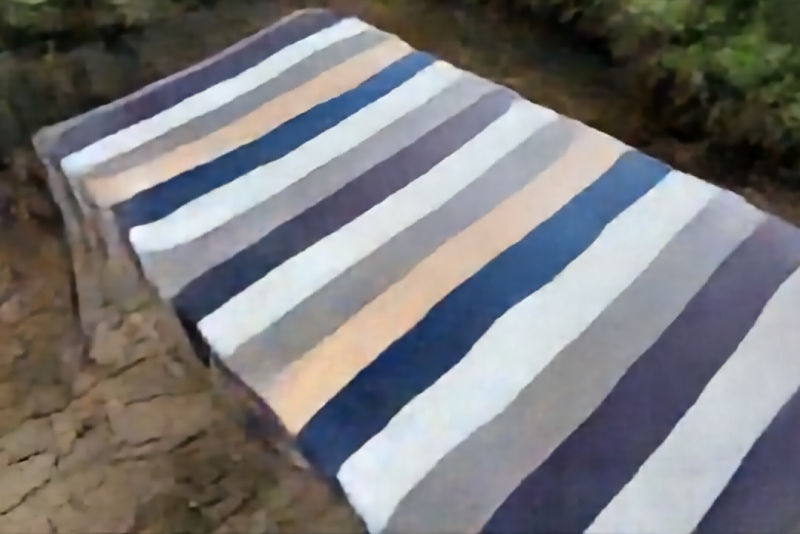} &
    \includegraphics[width=0.24\linewidth, valign=m]{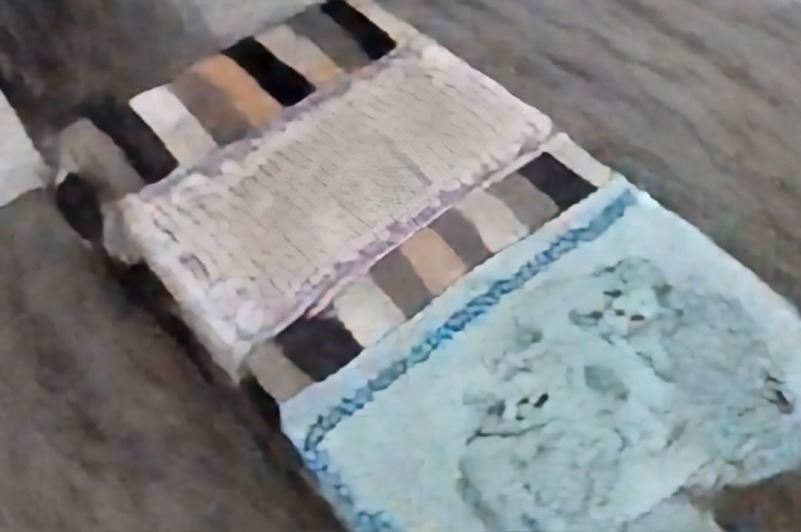} &
    \includegraphics[width=0.24\linewidth, valign=m]{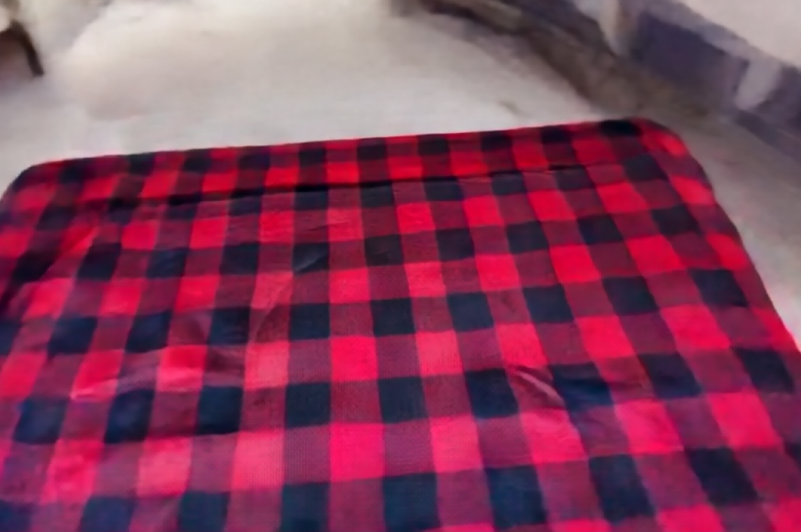} \\[37pt]
    \adjustbox{valign=m}{\rotatebox{90}{\large Ours}} &
    \includegraphics[width=0.24\linewidth, valign=m]{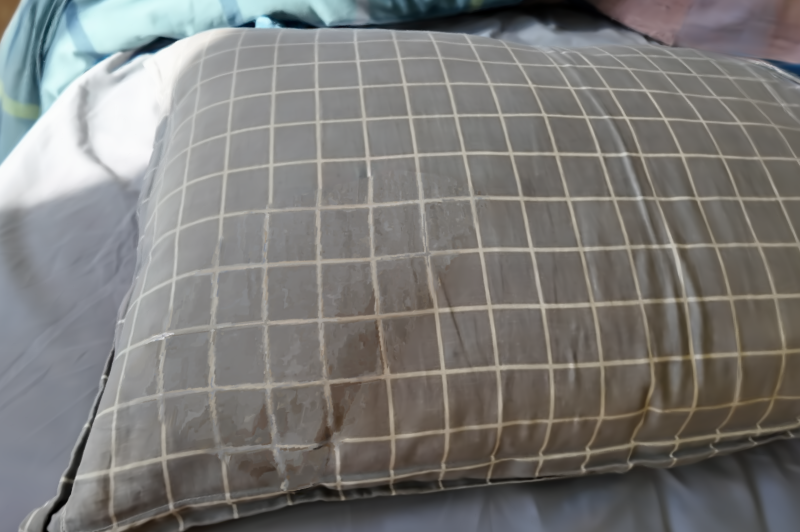} &
    \includegraphics[width=0.24\linewidth, valign=m]{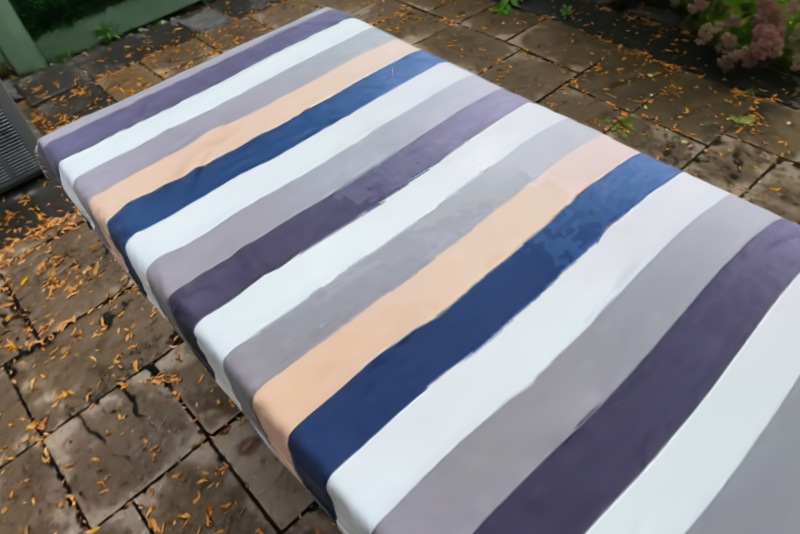} &
    \includegraphics[width=0.24\linewidth, valign=m]{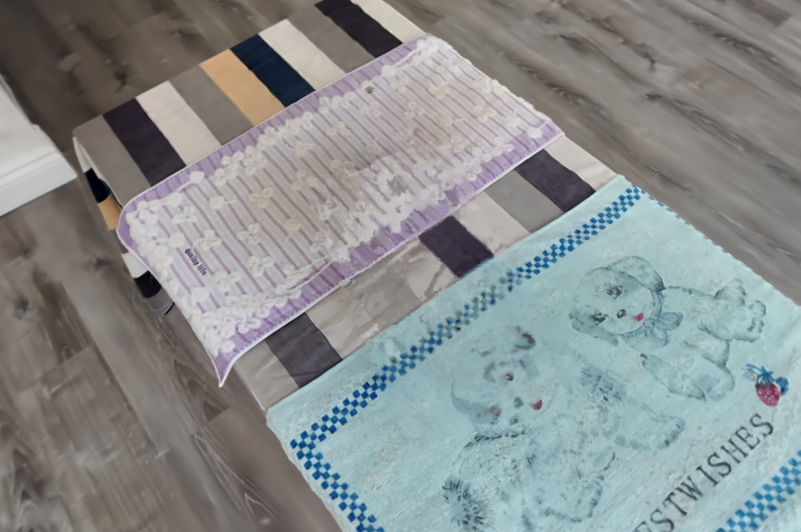} &
    \includegraphics[width=0.24\linewidth, valign=m]{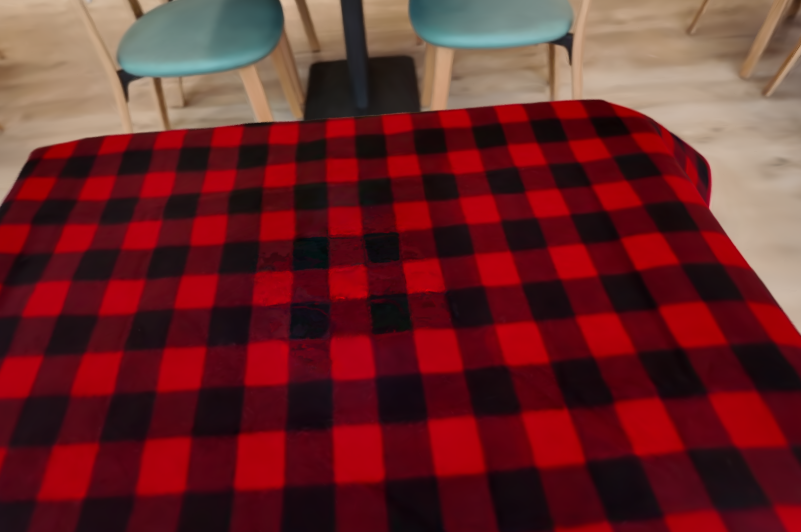} \\[37pt]
    \adjustbox{valign=m}{\rotatebox{90}{\large Ground truth}} &
    \includegraphics[width=0.24\linewidth, valign=m]{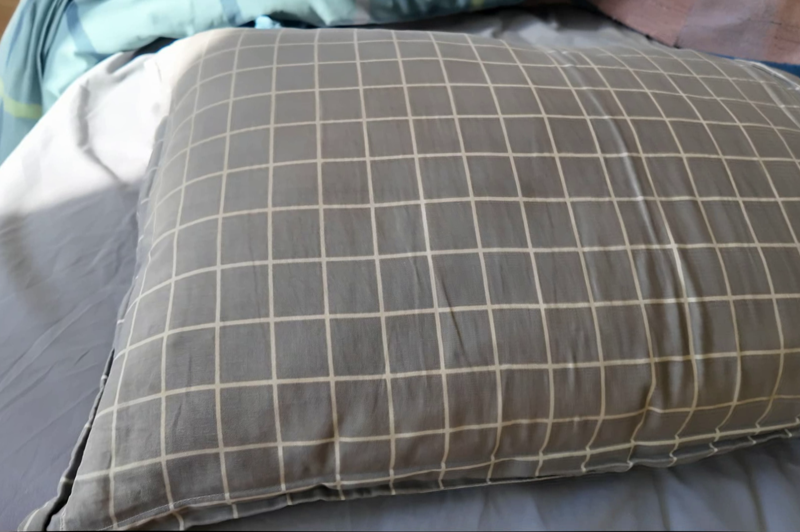} &
    \includegraphics[width=0.24\linewidth, valign=m]{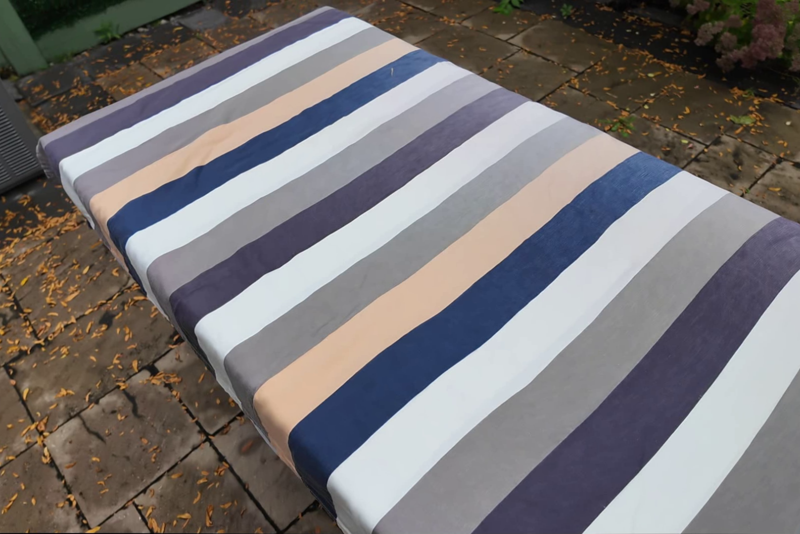} &
    \includegraphics[width=0.24\linewidth, valign=m]{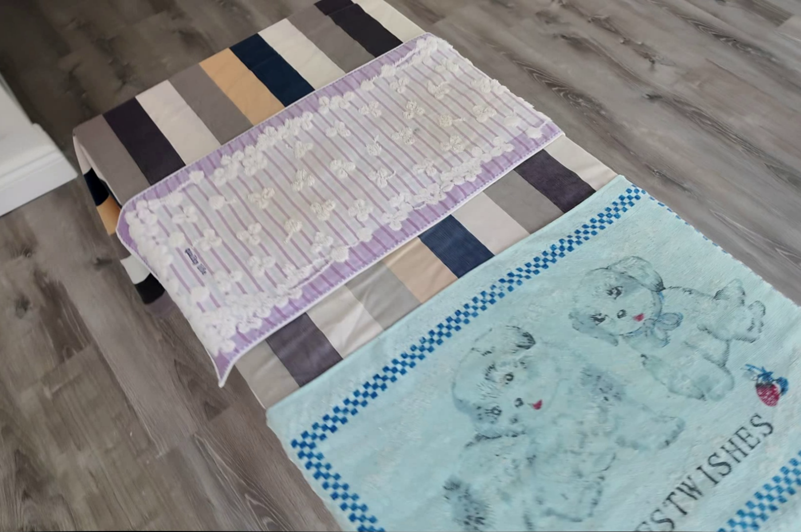} &
    \includegraphics[width=0.24\linewidth, valign=m]{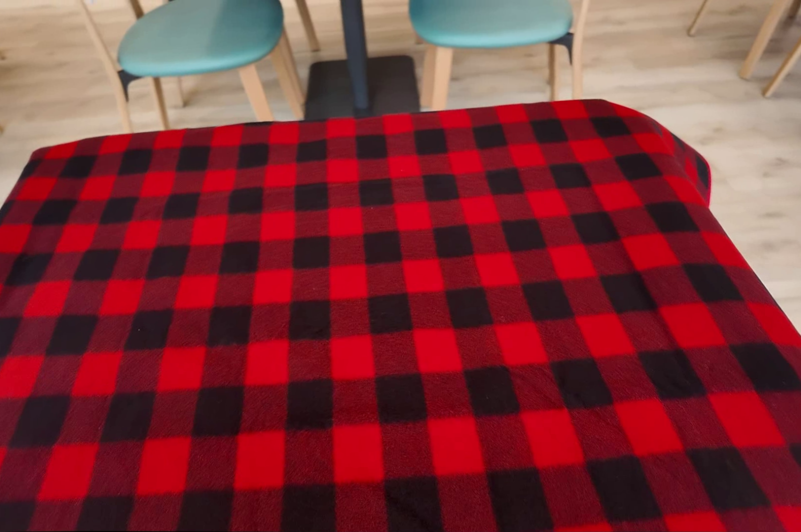} \\
 
\end{tabular}
  \caption{Qualitative comparisons with GEN3C \cite{ren2025gen3c}. Since GEN3C uses only sparse views to initiate the temporal inpainting frame, the surrounding environment is often poorly reconstructed. To ensure a fair comparison, we mitigate this issue by evaluating the results exclusively within the object mask, as shown in Figure \ref{table:gen3c}.}
  \label{fig/gen3c.png}
\end{figure*}

\end{document}